\documentclass{article}
    \PassOptionsToPackage{numbers, compress}{natbib}

\usepackage[utf8]{inputenc} % allow utf-8 input
\usepackage[T1]{fontenc}    % use 8-bit T1 fonts
\usepackage{geometry}
\usepackage{amsmath, amssymb, amsthm}
\usepackage[colorlinks,citecolor=blue,urlcolor=black,linkcolor=blue,pdfborder={0 0 0}]{hyperref}       % hyperlinks
 \usepackage[capitalize]{cleveref}  
\usepackage{url}            % simple URL typesetting
\usepackage{booktabs}       % professional-quality tables
\usepackage{amsfonts}       % blackboard math symbols
\usepackage{microtype}      % microtypography
\usepackage{xcolor}         % colors
\usepackage{natbib}
\usepackage{forloop}
\usepackage{graphicx}
\usepackage{multirow}
\usepackage{wrapfig}
\usepackage[font=small,labelfont=bf]{caption}
\usepackage{bm}
\usepackage{amsfonts}
\usepackage{nicefrac}       % compact symbols for 1/2, etc.
\usepackage[linesnumbered, ruled, algo2e]{algorithm2e}
\usepackage{algorithm}
\usepackage{algorithmicx}
\usepackage{caption}
\usepackage{subcaption}
\usepackage{enumitem}

\def\norm#1{\|{#1}\|} % A norm with 1 argument
\newcommand{\twonorm}[1]{\norm{#1}_2}
\providecommand{\argmin}{\mathop\mathrm{arg min}}
\def\balign#1\ealign{\begin{align}#1\end{align}}
\def\baligns#1\ealigns{\begin{align*}#1\end{align*}}
\def\balignat#1\ealign{\begin{alignat}#1\end{alignat}}
\def\balignats#1\ealigns{\begin{alignat*}#1\end{alignat*}}
\def\bitemize#1\eitemize{\begin{itemize}#1\end{itemize}}
\def\benumerate#1\eenumerate{\begin{enumerate}#1\end{enumerate}}
\def\textsum{{\textstyle\sum}}
\newenvironment{talign*}
 {\csname align*\endcsname}
 {\endalign}
\newenvironment{talign}
 {\csname align\endcsname}
 {\endalign}
 \def\balignst#1\ealignst{\begin{talign*}#1\end{talign*}}
\def\balignt#1\ealignt{\begin{talign}#1\end{talign}}
\def\defeq{\triangleq}
\newcommand{\reg}[0]{\pi} 
\newcommand{\loss}[0]{\ell}

\newcommand{\estt}[0]{\hat{\theta}_n(\lambda)} 
\newtheorem{theorem}{Theorem}
\newtheorem{lemma}{Lemma}
\newtheorem{assumption}{Assumption}
\newtheorem{definition}{Definition}
\newtheorem{proposition}{Proposition}
\newtheorem{remark}{Remark}
\newcommand{\E}{\mathbb{E}}

\newcommand{\D}{\mathcal{D}}
\newcommand{\Ss}{S}

\title{Algorithms that Approximate Data Removal: New Results and Limitations}

\author{Vinith M. Suriyakumar, Ashia C. Wilson}

\date{%
    MIT EECS\\
    \texttt{\{vinithms, ashia07\}@mit.edu}
    \\[2ex]
    \today}

\begin{document}
%\title{An Efficient Adaptive Machine Unlearning Algorithm}
%\title{Algorithms that Approximate Data Deletion}
%\title{ An Online Algorithm for Complying with GDPR }
\maketitle
\begin{abstract}
    We study the problem of deleting user data from machine learning models trained using empirical risk minimization. Our focus is on learning algorithms which return the empirical risk minimizer and approximate unlearning algorithms that comply with deletion requests that come streaming minibatches. Leveraging the infintesimal jacknife, we develop an online unlearning algorithm that is both computationally and memory efficient. Unlike prior memory efficient unlearning algorithms, we target models that minimize objectives with non-smooth regularizers, such as the commonly used $\ell_1$, elastic net, or nuclear norm penalties. We also provide generalization, deletion capacity, and unlearning guarantees that are consistent with state of the art methods. Across a variety of benchmark datasets, our algorithm empirically improves upon the runtime of prior methods while maintaining the same memory requirements and test accuracy. Finally, we open a new direction of inquiry by proving that all approximate unlearning algorithms introduced so far fail to unlearn in problem settings where common hyperparameter tuning methods, such as cross-validation, have been used to select models.
\end{abstract}

\section{Introduction}
 The right to be forgotten (RtbF) is considered a fundamental human right in many legal systems~\cite{rosen2011right, mcgoldrick2013developments, bygrave2014right}. The proliferation of techniques that employ user data to do things such as training and validating machine learning across a variety of organizations has led to reconsideration of how to interpret RtbF. The European Union's General Data Protection Regulation (GDPR), California's Consumer Privacy Act (CCPA), and Canada's proposed Consumer Privacy Protection Act (CPPA) are all examples of new pieces of legislation which attempt to codify the RtbF by requiring companies and organizations to delete a user's data by request~\cite{voigt2017eu}. 
 
 But \textit{what does it mean to delete a user's data?} User data, for example, can be recovered from trained machine learning models~\cite{shokri2017membership} and hyperparameter tuning procedures~\cite{papernot2021hyperparameter}. This suggests that data deletion should require organizations to take action on the models and algorithms derived from the data as well. This interpretation is consistent with the Federal Trade Commission's recent action ordering companies to delete the data from users who deleted their account as well as the models and algorithms derived from the users' data~\cite{FTC}. Forcing organizations to comply with each deletion request by retraining, however, comes with potentially significant monetary and time costs. Thus, it is worth asking if these costs can be managed while still maintaining model performance.  %associated with obtaining modern machine learning models. %developed using large amounts of data ~\citep{sekhari2021remember}.  %In lieu of relearning the model from scratch, several works have proposed methods  %but left data deletion guidelines intentionally ambiguous. % across a variety of organizations . Several laws have been introduced, including the 
%that require dele . %Does deleting one's data mean initiating the entire learning process with a new dataset consisting of the deleted datapoint? In other words, is the problem of unlearning as computationally expensive as the problem of learning? For which ML pipelines should it be required that a user's data be deleted upon request?
%

The cost concerns associated with RtbF compliance have prompted several recent works which formalize and study the problem of unlearning~\cite{cao2015towards}. The aim of these works is to develop techniques which delete user data from models that are inexpensive in both computation and memory while maintaining reasonable performance.
Most unlearning algorithms proposed so far focus on unlearning models obtained via empirical risk minimization~\cite{guo2020certified,sekhari2021remember,bourtoule2021machine,neel2021descent,marchant2021hard}. However, several of these methods are memory intensive~\cite{bourtoule2021machine, gupta2021adaptive}, requiring organizations to store several states of the model while training. We build on recent works~\cite{guo2020certified, sekhari2021remember} which show that for models that approximately minimize convex and sufficiently smooth ERM objectives, a Newton update to the current model can be applied in order to approximately unlearn in certain data directions. We show a different Newton estimate based on the classical infinitesimal jacknife developed in the robust statistics literature ~\cite{jaeckel1972infinitesimal,efron1982jackknife} results in a more computationally efficient unlearning algorithm in the online setting. While the current legislation only requires companies to satisfy delete requests in 30 days there are scenarios where immediate (online) delete requests are necessary such as the UK Biobank~\cite{ginart2019making}. We also show how to utilize the proximal infinitesimal jackknife~\cite{wilson2020approximate} to unlearn in settings when the model was trained using an ERM objective that is not smooth. Finally, we present an important and concerning failure mode of all approximate unlearning algorithms. %to stimulate discussion on the question of 

%\textit{What does it mean to delete a user's data?}.  %showcase when minimizing a quadratic approximation of the objective function is sufficient. 

%We summarize our results in the following. 
%\paragraph{ Contributions.} 
Our principal contributions are three-fold. %We present an online algorithm to comply with deletion requests.  %in based on computing an influence function. %based on the infinitesimal jackknife
\begin{itemize}[leftmargin=15pt]
   \item  {\em A less expensive batch algorithm for data removal.} We develop an online unlearning algorithm to unlearn a sequence of $m$ datapoints from a model that approximately minimizes a smooth and convex ERM objective that requires $O(d^2)$ storage and has running time $O(md^2)$. % where $\omega \in [2,2.38]$ is the exponent of matrix multiplication, 
    To unlearn $m$ datapoints in an online manner (i.e. through several batch requests), the algorithms developed in prior works~(e.g. \cite{guo2020certified, sekhari2021remember}) requires a similar storage, but have a longer $O(md^3)$ running time. By avoiding the cost of computing and inverting a different Hessian at each deletion request we improve on the computational requirements of previous memory efficient unlearning algorithms, while maintaining the same generalization, unlearning and deletion capacity guarantees. We empirically demonstrate significant computational savings without sacrificing test set accuracy on multiple datasets.   %focuses on developing unlearning algorithms for the setting where data requests come all at once~\cite{sekhari2021remember,guo2020certified,bourtoule2021machine} under storage and computation constraints. When adapted to the online setting, many of these techniques become significantly more expensive as they require inverting a Hessian for each datapoint that is unlearned (or storing a Hessian for each datapoint). We address this issue by showing that a simple more cost effective infinitesimal jackknife estimate can be used while maintaining the unlearning, generalization and deletion capacity guarantee of previous methods.  %Handles the streaming setting which is more natural than the batch setting proposed by~\cite{sekhari2021remember}. Less computationally expensive in both time and memory then~\cite{guo2020certified, bourtoule2021machine, sekhari2021remember} Naturally defends against adaptive sequences of deletion requests which SISA does not
    %\item 
    \item {\em Accommodating non-smooth regularizers.} We provide a generalization of our unlearning algorithm based on the proximal Newton method that can be used to efficiently delete data from models that minimize objective functions with non-smooth penalties. %Using state-of-the-art proximal Newton packages such as GLMnet~\cite{ } and QUIC~\cite{ }, e show that it is possible to comply with deletion requests efficiency while maintaining 
    We provide state-of-the art unlearning, deletion capacity and generalization guarantees for our new unlearning method. We empirically demonstrate significant computational savings without sacrificing test set accuracy in predicting Warfarin dosages.  
    \item {\em Uncovering failure modes of unlearning algorithms.} Most data processing pipelines are not as simple as writing down an objective function and running an algorithm out-of-the-box which returns the approximate empirical risk minimizer. A more common practice is to tune hyperparameters using techniques such as cross-validation. 
    We reveal a fundamental limitation of most data removal processes developed so far when hyperparameter tuning takes place.  
\end{itemize}

\section{Related Work}
Given the increasing concerns around privacy of user data, %and %regulations such as the Right to Be Forgotten from the EU's GDPR,
recent research on \textit{machine unlearning} studies how we can efficiently delete datapoints used to train models without retraining from scratch. This work was first initiated by~\citet{cao2015towards} whose definition requires the outputs of an unlearning algorithm to be identical to the outputs of the model produced by retraining. Since then, several works have provided different definitions of machine unlearning which can be separated as exact unlearning~(\cite{bourtoule2021machine, thudi2021necessity, gupta2021adaptive, garg2020formalizing}) or approximate unlearning~(\cite{guo2020certified, sekhari2021remember, ginart2019making, dwork2014algorithmic, neel2021descent}). 

Our work is focused on satisfying approximate unlearning definitions inspired by differential privacy~\cite{dwork2014algorithmic} defined by~\cite{guo2020certified} as $(\epsilon, \delta)$-certified removal or by~\cite{sekhari2021remember} as $(\epsilon, \delta)$-unlearning. This definition requires the output distribution of the unlearning algorithm to be similar to that obtained by retraining from scratch using the original training set with the requested datapoints removed from it. We focus on this definition because algorithms for exact unlearning such as statistical query learning~\cite{cao2015towards}, SISA~\cite{bourtoule2021machine}, and its adaptive variant~\cite{gupta2021adaptive} have high computational and memory cost. 

Prior algorithms developed for approximate unlearning use a variety of techniques: perturbed gradient descent~\cite{neel2021descent, ullah2021machine}, Newton style updates~\cite{guo2020certified, sekhari2021remember}, and projected residual updates~\cite{izzo2021approximate}. The perturbed gradient descent and projected residual update methods provide theoretical error guarantees for the empirical training loss but fail to provide any generalization guarantees. In addition, while these methods reduce the computation burden associated with exact unlearning, their high memory costs are still similar to exact unlearning methods.

Given these issues, Newton update unlearning algorithms were proposed to efficiently satisfy approximate unlearning. Our work is closest to~\cite{guo2020certified, sekhari2021remember} who use Newton updates to efficiently delete data for generalized linear models.~\citet{sekhari2021remember} improve upon~\citet{guo2020certified} in multiple ways: (i) by not requiring full access to the training dataset, providing (ii) generalization guarantees, and (iii) removing the requirement for randomization in the learning algorithm itself (which often reduces utility). Yet the algorithm developed by~\cite{sekhari2021remember} targets the batch setting which is unrealistic in practice and they provide no empirical results demonstrating the efficacy of their algorithm. If implemented in an online way, their algorithm would suffer a much larger computational cost given the requirement to compute and invert a new Hessian for each delete request. Finally, neither of these algorithms provide theoretical guarantees for the commonly used models that are obtained from objective functions with non-smooth penalties.  

We address several of these issues using the (proximal) infinitesimal jacknife~\cite{giordano2019swiss, jaeckel1972infinitesimal, efron1982jackknife}. Our algorithm can handle online delete requests selected in an adaptively adversarial manner (a failure mode of many previous algorithms pointed out by~\citet{gupta2021adaptive}) making it more practical for real world use. Furthermore, we provide generalization and deletion capacity guarantees similar to~\citet{sekhari2021remember} for both smooth and non-smooth regularizers. To do so, we leverage similar guarantees found in approximate cross-validation literature~\cite{wilson2020approximate}. Finally, inspired by recent working showing that hyperparameter tuning can leak user data~\cite{papernot2021hyperparameter} we demonstrate that all approximate unlearning algorithms introduced so far fail to unlearn in settings where hyperparameter tuning has taken place to choose a model. 

\section{Methods and Results}
\label{sec:methods_results}
\paragraph{Learning} Consider the objective function 
%\begin{align*}
%F(z, \theta, \lambda)  =  \ell (z, \theta) + \lambda \pi(\theta) 
%\end{align*}
comprised of a loss function $\ell$, a regularizer $\pi$ and regularization parameter $\lambda \in \Lambda \subseteq [0,\infty]$. %Let $\mathbb{P}$ represents the measure associated with the data distribution $\mathcal{D}$ over the instance space $\mathcal{Z}$. 
The goal of learning is to find a parameter $\theta^\ast(\lambda)$ which minimizes the population risk
\begin{align*}
   %\theta^\ast(\lambda) \triangleq %\begin{cases}
   %\text{argmin}_{\theta}\,\,\{
   F(z, \theta, \lambda) \triangleq \ \E_{z \sim \mathcal{D}} [\ell (z, \theta)] + \lambda \pi(\theta)% \} %& \lambda \in [0,\infty) %\\ \text{argmin}_{\theta} \,\,\pi(\theta)& \lambda = \infty
   %\end{cases}
\end{align*}
%Here, we overload notation and write $\ell(\mu , \theta)\triangleq  \int \ell(\mu , \theta) d \mu (z)$  so that $\ell(\mathbb{P}, \theta) \triangleq \mathbb{E}_{z \sim \mathcal{D}} [\ell(z, \theta)]$.
Given the distribution $\mathcal{D}$ is often inaccessible, practitioners often instead find a model $ \hat{\theta}_n(\lambda)$ which (at least approximately) minimizes the empirical risk
\begin{talign}\label{eq:erm}
 F_n(z, \theta, \lambda)   %\begin{cases} 
   \triangleq  \tfrac{1}{n} \sum_{i=1}^n \ell(z_i, \theta) + \lambda \pi(\theta,) % & \lambda \in [0,\infty) \\ \text{argmin}_{\theta} \,\,\pi(\theta)& \lambda = \infty
   %\end{cases}
\end{talign}
corresponding to a given dataset $\Ss= (z_1, \dots z_n)$.%\footnote{Here, we overload notation and use $\Ss$ to represent a set of datapoints as well as indices corresponding to those datapoints}
\paragraph{Unlearning} Having used a dataset $\Ss$ to train and publish a model $\hat{\theta}_n(\lambda)$, a set of $m$ users in the training set $U\subset \Ss$ might request that their datapoints be deleted and that any models produced using their data be removed. To comply with this request, an organization might find the minimizer  $\theta_{n, -U}(\lambda)$ of the leave-$U$-out objective 
\balignt\label{eq:cvest}
%\hat{\theta}_{n,-i}(\lambda) 
%= \argmin_\beta   \tfrac{1}{n-1}\textsum_{j \neq i} \loss(X_j,\beta) + \frac{n}{n-1} \lambda\reg(\beta) 
%&= \argmin_\theta 
F_{n, -U}(z, \theta, \lambda)  \triangleq \tfrac{1}{n-m}\textsum_{z \in \Ss\backslash U} \ell (z,\theta) + \lambda\reg(\theta). 
\ealignt

While {\em reoptiminzing the leave-one-out objective from scratch constitutes a baseline for the problem of unlearning}, the computational cost makes complying with every data delete request in this way undesirable.
Training from scratch satisfies the notion of unlearning~\citep{sekhari2021remember} formalized in~\cref{def:unlearn}.

\begin{definition}[$(\epsilon,\delta)$-unlearning \citep{sekhari2021remember}]\label{def:unlearn}
Let $\Ss$ be a fixed training set and $A: \Ss \rightarrow \theta$ be an algorithm that trains on  $\Ss$ and outputs a model $\theta \in \Theta$. For an $\epsilon>0$ and set of delete requests $U \subseteq S$, we say that a removal mechanism $M$ is $(\epsilon, \delta)$-unlearning for learning algorithm $A$ if $\forall\, \mathcal{T} \subseteq {\Theta}$ and $\Ss \subseteq \mathcal{Z}$, the following two conditions are satisfied:

\begin{align*}
 P(M(U, A(\Ss), T(S))\in \mathcal{T})&\leq e^{\epsilon} P(M(\emptyset, A(\Ss \backslash U), T(S \backslash U))\in \mathcal{T}) + \delta, \quad \text{and}\\
P(M(\emptyset, A(\Ss \backslash U), T(S \backslash U))\in \mathcal{T})&\leq e^{\epsilon} P(M(U, A(\Ss), T(S))\in \mathcal{T}) + \delta
\end{align*}

\end{definition}

Finally, we point out that like most previously proposed unlearning algorithms, we do not require $T(S)$ to contain the entire training set, but instead propose unlearning algorithms for which $T(S)$ is independent of $n$.  
%\clearpage
\subsection{Unlearning models obtained via regularized empirical risk minimization} We recommend use of the following proximal operator to comply with delete requests
\begin{align}\label{eq:prox-operator}
{\bf \text{ prox}}_{\lambda\pi}^{H}(v)\defeq \argmin_{\theta \in \mathbb{R}^d} \frac{1}{2} \| v- \theta\|_{H}^2 + \lambda \pi(\theta)
\end{align}
which allows us to handle objective functions that incorporate non-smooth regularizers, such as the $\ell_1$, elastic net or nuclear norm penalty. More specifically, having deleted the datapoints in the set $U$, we propose~\cref{algo:1} to delete the data of an additional user $j$. 
\begin{algorithm}[ht!]
\begin{algorithmic}
\State {\bf Input:} $|U| = m$, Hessian of  loss or objective, $\tilde{\theta}_{n,-U}(\lambda)$, $\bar{\theta}^N_{n,-U}(\lambda)$ $\hat{\theta}_{n}(\lambda)$, and delete request $j$
\State {\bf Set:} $H _\ell= \frac{1}{n}\sum_{i=1}^n\nabla^2_{\theta}\loss(z_i,\hat{\theta}_n)$, $H _f= \frac{1}{n}\sum_{i=1}^n\nabla^2_{\theta}f(z_i,\hat{\theta}_n,\lambda)$, $\bar{\theta}_{n, -\emptyset}^N(\lambda) = \tilde{\theta}_{n, -\emptyset}(\lambda) = \estt$
\State {\bf If $\pi$ is not smooth:}
\State {}\quad \quad {\bf Compute:}
\begin{subequations}
\begin{talign}
%\bar{\theta}_{n, -(U\cup \{j\})}^N(\lambda) &=  \\
\bar{\theta}_{n, -(U\cup \{j\})}(\lambda) &= {\bf \text{ prox}}_{\lambda\pi}^{H_\ell}\Big( \bar{\theta}_{n, -U}^N(\lambda) + \frac{1}{n}H_\ell^{-1}\nabla \ell(z_j, \hat{\theta}_{n}(\lambda)  )\Big)\\
c&= \frac{(m+1)^2(2CL\mu + ML)}{\mu^2n^2}\frac{\sqrt{2\ln(1.25/\delta)}}{\epsilon}
%&= \argmin_{\theta\in\reals^d}\Big(\half \|\est_n(\lambda) - \theta\|_{H_\loss}^2 + \frac{1}{n}\theta ^\top \nabla \ell(z_j, \bar{\theta}_{n}(\lambda)  )+ \lambda \pi(\theta)\Big)
\end{talign}
\end{subequations}
\State {}\quad \quad {\bf Store: }$\bar{\theta}_{n, -(U\cup \{j\})}^N(\lambda) = \bar{\theta}_{n, -U}^N(\lambda) + \frac{1}{n}H_\ell^{-1}\nabla \ell(z_j, \hat{\theta}_{n}(\lambda)  )$\\
\State{\bf Else compute:} 
%\State {}\quad \quad{\bf Compute:}
\begin{subequations}
\begin{talign}
\bar{\theta}_{n, -(U\cup \{j\})}(\lambda) &= \tilde{\theta}_{n, -U}(\lambda) + \frac{1}{n}H_f^{-1}\nabla \ell(z_j, \hat{\theta}_{n}(\lambda)  )\\
c& = \frac{(2m+1)(2CL\mu + ML)}{\mu^2n^2}\frac{\sqrt{2\ln(1.25/\delta)}}{\epsilon} %\hat{\theta}_{n, -m,-j}(\lambda) &= \hat{\theta}_{n, -i}(\lambda) + \frac{1}{n-1} H^{-1} \nabla \ell(z_j, \theta_{n}(\lambda)) + \sigma
% &= \argmin_{\theta\in\reals^d}\Big(\half \|\est_n(\lambda) + \sigma - \theta\|_{H_\loss}^2 + \frac{1}{n-m}\theta ^\top \nabla \ell(z_j, \hat{\theta}_{n}(\lambda)  )+ \lambda \pi(\theta)\Big)\\
\end{talign}
\end{subequations}
\State {\bf Sample:} $\sigma \sim \mathcal{N}(0, cI)$ \\
{\bf Return:} $\tilde{\theta}_{n, -(U\cup \{j\})}(\lambda)  := \bar{\theta}_{n, -(U\cup \{j\})}(\lambda) + \sigma$\\
%  \begin{talign*}
% %\text{or}\\
% \tilde{\theta}_{n, -m,-j}(\lambda) &= {\bf \text{ prox}}_{\lambda\pi}^{H_\ell}\Big(\hat{\theta}_{n, -m}(\lambda) - \frac{1}{n-m}H_\ell^{-1}\nabla \ell(z_j, \hat{\theta}_{n,-U}(\lambda)(\lambda)  ) + \sigma \Big) \\%\hat{\theta}_{n, -m,-j}(\lambda) &= \hat{\theta}_{n, -i}(\lambda) + \frac{1}{n-1} H^{-1} \nabla \ell(z_j, \theta_{n}(\lambda)) + \sigma
% &= \argmin_{\theta\in\reals^d}\Big(\half \|\est_n(\lambda) + \sigma - \theta\|_{H_\loss}^2 + \frac{1}{n-m}\theta ^\top \nabla \ell(z_j, \hat{\theta}_{n}(\lambda)  )+ \lambda \pi(\theta)\Big)\\
% \end{talign*}
\end{algorithmic}
\caption{Infinitesimal Jacknife (\textbf{IJ}) Online Unlearning Algorithm}
\label{algo:1}
\end{algorithm}

%\paragraph{Memory and Computation} ~\cref{algo:1} entails calculating and inverting the full Hessian $H_f$ or $H_\ell$ and storing it (memory cost $O(d^2)$ and computation cost $O(d^3)$). It also entails storing the full data model $\estt$. While the computation cost is dominated by the inversion of the Hessian and the cost of the proximal step when $\pi$ is not smooth, we can leverage linear algebra libraries and GPUs to make this computation more efficient. In addition, both algorithms require storing the current model $\tilde{\theta}_{n, -U}(\lambda)$, and in the setting where $\pi$ is not smooth $\bar{\theta}_{n,-U}(\lambda)$ must be stored. During runtime,~\cref{algo:1} requires matrix-vector multiplication and vector addition (as well as proximal step in the non-smooth setting).

%In this section, we provide unlearning algorithms for objective functions and loss functions that are convex. 

\paragraph{Guarantees}
While we give guarantees for functions that are strongly convex, we rely on standard reductions from the convex setting to the strongly convex setting (i.e. based on defining a objective function $ F(\theta) = \tilde{F}(\theta) + (\mu/2)\|\theta\|^2$ when $\tilde{F}$ is convex). Furthermore, similar to~\citet[3.3]{guo2020certified} our methods can be used for unlearning in the non-convex setting when the deep learning model applies a simple convex model to a differentially private feature extractor~\cite{abadi2016deep}.  % assumptions we need and how they relate to previous work
\begin{assumption}[Smooth regularizer]\label{ass:1}
For any $z \in \mathcal{Z}$, the objective function $F(\theta, z,\lambda)$ is $\mu$-strongly convex and $L$-Lipschitz with $M$-smooth Hessian. The loss has $C$-Lipschitz Hessians. %In addition, suppose $f$ satisfies has Lipscitz Hessians. 
%\begin{talign*}
%B^U = \frac{1}{m}\sup_{\lambda \in \Lambda} \frac{1}{n}\sum_{i\in U} \text{\em Lip}(\nabla_\theta \ell(z_i,\cdot))\|\nabla_\theta \ell(z_i, \est_n(\lambda))\| <\infty.
%\end{talign*}
\end{assumption}
\begin{assumption}[Non-smooth regularizer]\label{ass:2}
For any $z \in \mathcal{Z}$, the loss function $\loss(\theta, z)$ is $\mu$-strongly convex and $L$-Lipschitz with $M$-smooth and $C$-Lipschitz Hessians. The regularizer $\pi(\theta)$ is convex. %In addition, suppose $f$ satisfies has Lipscitz Hessians. 
%\begin{talign*}
%B^U = \frac{1}{m}\sup_{\lambda \in \Lambda} \frac{1}{n}\sum_{i\in U} \text{\em Lip}(\nabla_\theta \ell(z_i,\cdot))\|\nabla_\theta \ell(z_i, \est_n(\lambda))\| <\infty.
%\end{talign*}
\end{assumption}
With either of these assumptions, it's possible to show that the denoised output of~\cref{algo:1} is $O(m^2/n^2)$ close to the exact unlearned model. We formalize this proximity result in the following~\cref{lemma:noise}.

\begin{lemma}[Proximity to the baseline estimator]\label{lemma:noise}
Suppose $F$ satisfies Assumption~\ref{ass:1} or Assumption~\ref{ass:2}. Let $\Ss$ be the dataset of size $n$ sampled from $\mathcal{D}$ and $U \subseteq \Ss$ denote the set of $m$ delete requests. Consider $\bar{\theta}_{n,-U}(\lambda)$, i.e. the output of~\cref{algo:1} without noise term $\sigma$ added and the model $\hat{\theta}_{n,-U}(\lambda)$ obtained by minimizing the leave-U-out objective $F_{n,-U}$. Then,
\begin{subequations}
\begin{talign}
\twonorm{ \hat{\theta}_n(\lambda)- \hat{\theta}_{n, -U}(\lambda)} 
 &\leq \frac{m L}{\mu n}, \quad \quad \text{\em and}\\
% \end{align}
%\begin{align}
\label{eq:ineq1}\|\hat{\theta}_{n,-U}(\lambda) - \bar{\theta}_{n, -U}(\lambda) \|_2 &\leq \frac{2m^2 CL}{n^2\mu^2 } + \frac{m^2 ML^2}{n^2 \mu^3}  .
\end{talign}
\end{subequations}
\end{lemma}
%This helps guide how to choose the noise term as we detail in~\cref{App: }
\cref{lemma:noise} implies adding the noise term $\sigma \propto O(m^2/n^2)$ will result in the desired unlearning guarantee. 
The following~\cref{thm:1} outlines this guarantee as well as the proximity of the unlearnt model to the test loss minimizer.  The proof is contained in Appendix~\ref{app:closeness_proof}.
\begin{theorem}[Unlearning and generalization guarantees]\label{thm:1} Suppose the loss function satisfies Assumption~\ref{ass:1} or~Assumption~\ref{ass:2} and consider any learning algorithm that returns a model $O(1/n^2)$ close to the empirical risk minimizer $\hat{\theta}_n(\lambda)$ trained on any dataset $\Ss \sim \mathcal{D}$ of size $n$. Then %Let $c = $ when $\pi$ is not smooth and $c = $ when $\pi$ is smooth. Then:
%\begin{itemize}
    %\item T
    the output $\tilde{\theta}_{n,-U}(\lambda)$ of~\cref{algo:1}, where $|U| = m$, satisfies the test error bound
    \begin{talign*}\mathbb{E}[F(\tilde{\theta}_{n, -U}(\lambda)) -  F(\theta^\ast(\lambda))] \leq O\left( \frac{(2m^2CL^2\mu + ML^3m^2}{\mu^3n^2}\cdot\frac{\sqrt{d}\sqrt{ln(1/\delta)}}{\epsilon} + \frac{4mL^2}{\mu n}\right).\end{talign*}
    Furthermore, the unlearning~\cref{algo:1} results in $(\epsilon, \delta)$-certifiable removal of $\forall {\bf z} \in U\subseteq \Ss$. 
%\end{itemize}
\end{theorem}

\subsubsection{Deletion Capacity}

\citet{sekhari2021remember} introduce the notion of deletion capacity which formalizes how many samples can be deleted from a the model parameterized by the original empirical risk minimizer $\estt$ while maintaining reasonable guarantees on the test loss. We restate the definition here.
    \begin{definition}[Deletion capacity~\cite{sekhari2021remember}]Let $\epsilon,\delta,\gamma>0$ and $\Ss$ be a dataset of size $n$ drawn i.i.d From $\D$, and let $F(\theta,z,\lambda)$ be an objective function. For a learning algorithm and removal mechanism $M$ that satisfies $(\epsilon,\delta)$-unlearning of all ${\bf z} \in U$, where $U$ is the set of delete requests, the deletion capacity $m_{\epsilon,\delta,\gamma}^{A,M}(d,n)$ is defined as the maximum number of samples that can be unlearned while still ensuring the excess population risk is $\gamma$. Specifically,
    \begin{talign*}
    m_{\epsilon,\delta,\gamma}^{A,M}(d,n) \defeq \max \left\{m \,|\, \mathbb{E}\left[\max_{U\subseteq \Ss: |U| \leq m} F(M(A(\Ss),\Ss, U) - F(\theta(\lambda)^\ast) \right] \leq \gamma\right\}
    \end{talign*}
where the expectation is with respect to $\Ss \sim \D$ and output of the learning algorithm 
$A$ and removal mechanism $M$.
\end{definition}

\citet{sekhari2021remember} provide both an upper bound and lower bound on the deletion capacity of unlearning algorithms. We recount both bounds and show our algorithm achieves the same bounds.

\begin{theorem}[Deletion capacity upper bound~\citep{sekhari2021remember}]
Let $\delta \leq 0.005$ and $\epsilon = 1$. There exists a 4-Lipschitz and 1-strongly convex loss function f, and a distribution $\mathcal{D}$ such that for any learning algorithm $A$ and removal mechanism $M$ that satisfies $(\epsilon, \delta)$-unlearning and has access to all undeleted samples $S \backslash U$, then the deletion capacity is:
\begin{align*}
m_{\epsilon,\delta,\gamma}^{A,M}(d,n) \leq c n,
\end{align*}
where the constant $c$ depends on the Lipschitz constants $L$, $M$, strongly convex constant $\mu$, and boundedness constant $C$ from Assumptions~\ref{ass:1} or~\ref{ass:2} and $c < 1$.
    
\end{theorem}

\begin{theorem}[Deletion capacity lower bound~\citep{sekhari2021remember}] \label{thm:del_cap_lower}
    Let $\epsilon, \delta > 0$ and $\gamma = 0.01$, $\Ss$ be a dataset of size $n$ drawn i.i.d from $\D$, and  $F(\theta,z,\lambda)$ be an objective function satisfying Assumption~\ref{ass:1} or Assumptions~\ref{ass:2}. Consider a learning algorithm $A$ that returns the empirical risk minimizer and unlearning algorithm $M$. Then the deletion capacity is:
    \begin{talign}\label{eq:lb}
    m_{\epsilon,\delta, 0.01}^{A,M}(d,n) \geq c \cdot \frac{n \sqrt{\epsilon}}{(d \text{\em log}(1/\delta))^{\frac{1}{4}}}
    \end{talign}
    where the constant $c$ depends on the Lipschitz constants $L$, $C$, and $M$.
    \end{theorem}

%DP for dataset distances of $m$ by definition unlearns $m$ samples.
\begin{theorem}[Deletion capacity from unlearning via DP~\citep{sekhari2021remember}] There exists a polynomial time learning algorithm $A$ and removal mechanism $M$ of the form $M(U, A(S), T(S)) = A(S)$ such that the deletion capacity is:
\begin{talign*}
 m_{\epsilon,\delta, 0.01}^{A,M}(d,n) \geq \tilde{\Omega}\left(\frac{n\epsilon}{\sqrt{d\log(e^{\epsilon}/\delta})}\right)
\end{talign*}
where the constant in $\Omega$-notation depends on the properties of the loss function $f$.
    
\end{theorem}
As an extensions of a result developed by~\citet[Section C]{bassily2019private}, \citet{sekhari2021remember} show that any unlearning algorithm that ignores the samples $U$ can't improve upon the DP lower bound. This motivates the study of algorithms specifically designed for unlearning which leverage samples from $U$ instead of algorithms based on DP.
In~\cref{app:del_cap}, we provide details showing that our Algorithm~\ref{algo:1} achieves the lower bound~\eqref{eq:lb}, where the constant $c$ depends on the Lipschitz constants $L$, $M$, strongly convex constant $\mu$, and boundedness constant $C$. 
% \begin{theorem}[Deletion capacity lower bound of Algorithm~\ref{algo:1}]\label{thm:del_cap}
%     Let $\epsilon, \delta, \gamma > 0$, and $\Ss$ be a dataset of size $n$ drawn i.i.d From $\D$, and let $F(\theta,z,\lambda)$ be an objective function. Given a learning algorithm and unlearning algorithm $M$ (Algorithm~\ref{algo:1}) such that the objective function $F(\theta, z, \lambda)$ satisfies Assumption~\ref{ass:1} or~\ref{ass:2} then the deletion capacity is:
%     \begin{talign*}
%     m_{\epsilon,\delta, \gamma}^{A,M}(d,n) \geq c \cdot \frac{n \sqrt{\epsilon}}{(d \text{log}(1/\delta))^{\frac{1}{4}}}
%     \end{talign*}
%     where the constant $c$ depends on the Lipschitz constants $L$, $M$, strongly convex constant $\mu$, and boundedness constant $C$.
%     \end{theorem}
%  Our deletion capacity is consistent with~\cite{sekhari2021remember} except for a difference in constant $c$. We explicitly derive $c$ in our proof of this theorem in Appendix~\ref{app:}.

\paragraph{Comparison to previous results}
We compare our method to~\citet{sekhari2021remember} and~\citet{guo2020certified} who propose optimizing a second-order Taylor approximation (\textbf{TA}) to the leave-$U$-out objective function~\eqref{eq:cvest} for batch removal of $U$. This results in the Newton-like removal mechanism~\eqref{eq:s}.
\begin{talign}\label{eq:s}
\tilde{\theta}_{n,-U}(\lambda) = \estt + \frac{1}{n-m}\left(\frac{1}{n-m}\sum_{z \in\Ss \backslash U}\nabla^2_\theta F(z,\estt,\lambda)\right)^{-1}  \sum_{z\in U}\nabla\ell(z,\estt). 
\end{talign}
 
{\em Comparison of Assumptions. } On the one hand, our Assumption~\ref{ass:1} is slightly more restrictive than that of~\citet{sekhari2021remember} given we require boundedness of the Hessian loss to obtain our unlearning, generalization and deletion capacity guarantees. Notably, this assumption does not rule out any of the most common convex objective functions (e.g. least squares, logistic, hinge, or cross entropy loss functions). On the other hand, our technique allows for non-smooth regularizers which are common in modern machine learning pipelines.

{\em Comparison of Computation.} 
~\cref{algo:1} entails calculating and inverting the full Hessian $H_f$ or $H_\ell$ and storing it (memory cost $O(d^2)$ and computation cost $O(d^3)$). It also entails storing the gradient of each point evaluated on the full data model $\estt$, as well as the current model $\tilde{\theta}_{n, -U}(\lambda)$. In the setting where $\pi$ is not smooth, $\bar{\theta}_{n,-U}(\lambda)$ must be stored. During runtime,~\cref{algo:1} requires matrix-vector multiplication and vector addition (as well as proximal step in the non-smooth setting).
For unlearning in settings where deletion requests come in a online manner, our technique is computationally more efficient. This is because removal mechanism~\eqref{eq:s} requires computing and inverting a different Hessian that depends on the user requesting the deletion. Therefore, outside simple settings, complying with such requests involves a computational cost of $O(md^3)$ to remove $m$ datapoints. 
 
 {\em Comparison of unlearning, generalization and deletion capacity.} As summarized above, the generalization, unlearning and deletion capacity results for our removal mechanism are essentially equivalent to the unlearning, generalization, and deletion capacity results of~\cite{sekhari2021remember}. 
  %  \item {\em Deletion capacity}
%\end{itemize}
\begin{remark}[Extending~\citet{sekhari2021remember} to non-smooth regularizers] Following from the equivalence we show between the batch and online setting in Appendix~\ref{app:batch_eq_stream}, we can extend our use of the proximal operator to~\eqref{eq:s} which would extend the results of~\citet{sekhari2021remember} to non-smooth regularizers. Similar unlearning, deletion capacity and generalization results are obtained. We provide details in~\cref{app:ext}. 
\end{remark}
\section{Experiments}
In this section we will refer to retraining from scratch as \textbf{RT}, the Algorithm~\eqref{eq:s} as \textbf{TA} and our \cref{algo:1} as \textbf{IJ}.
We empirically demonstrate the benefits  of our algorithm over both \textbf{RT} and \textbf{TA} in three different settings: (i) smooth regularizers where we train a logistic regression model with an $\ell_2$ penalty to predict between the digit 3 and 8 from the MNIST dataset~\cite{lecun1998mnist}, (ii) non-smooth regularizers where we train a logistic regression model with an $\ell_1$ penalty to predict whether an individual was prescribed a Warfarin dosage of $> 30$ mg/week from a dataset released by the International Warfarin
Pharmacogenetics Consortium~\cite{international2009estimation}, and (iii) non-convex training where we apply a logistic regression model with an $\ell_2$ penalty to predict street digits signs 3 and 8 from SVHN~\cite{netzer2011reading} on representations extracted from a differentially private feature extractor with $\epsilon = 0.1$ (similar to the setup of~\citet{guo2020certified}). For all algorithms, we tune $\lambda$ over the set $\{10^{-3}, 10^{-4}, 10^{-5}, 10^{-6}\}$. In the Appendix~\ref{app:datasets}, we provide further details about each dataset and the code to produce our models is attached separately.

 We present results in the main paper for $\lambda = 1^{-3}$ and provide other results in Appendix~\ref{app:exps}. All models are trained on a single NVIDIA Tesla T4 GPU and 16 2.10GHz Xeon(R) Silver 4110 CPU cores. We focus our evaluation on total runtime in seconds and test set accuracy as the number of delete requests increases. Given that the current window for complying with delete requests for GDPR is one month, the runtime savings on our plots occur at month X where X is provided by the y-axis. We note that the "Right to Be Forgotten" is a much broader right and our experimental findings demonstrate that this can be satisfied much more efficiently for convex problems.  For all results, we provide the average over three different runs and provide standard error bars.
% We empirically demonstrate the benefits  of our algorithm over both \textbf{RT} and \textbf{Taylor Approximation} in two settings: (1) training a logistic regression model with an $\ell_2$ penalty to predict between the digit 3 and 8 from the MNIST dataset~\cite{lecun1998mnist} and (2) training a logistic regression model with an $\ell_1$ penalty (non-smooth regularizer) to predict warfarin dosing~\cite{international2009estimation}. We tune the strength of regularization $\lambda$ over the following set $=\{1e^{-3}, 1e^{-4}, 1e^{-5}, 1e^{-6}\}$. All models are trained on a single NVIDIA V100 GPU and 16 2.10GHz Xeon(R) Silver 4110 CPU cores. For both \textbf{IJ}  and\textbf{TA} we fix the standard deviation of the Gaussian noise distribution at $c = 0.1$. 

\begin{figure}[t!]
\centering
\begin{subfigure}{.4\textwidth}
  \centering
  \includegraphics[width=\linewidth]{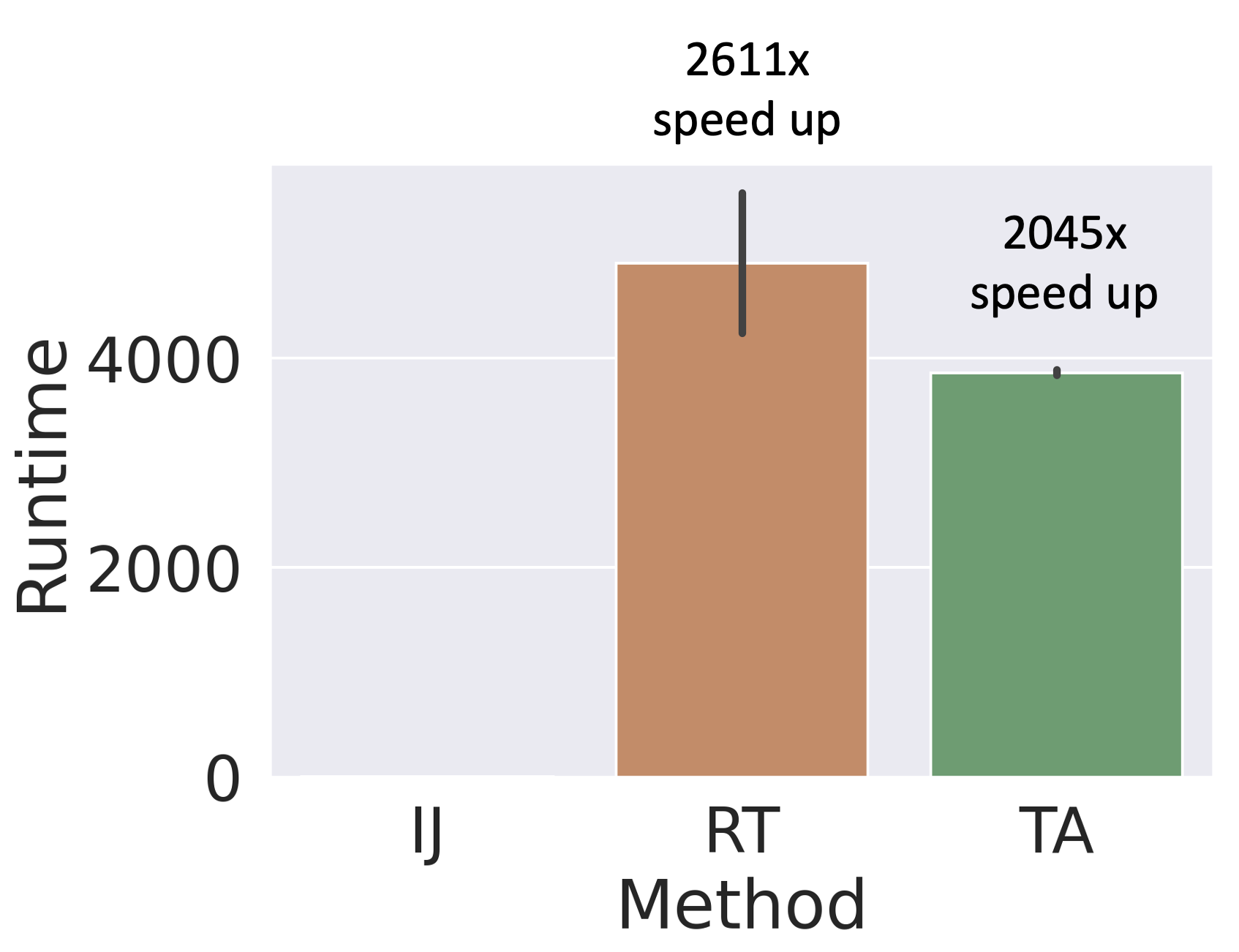}
  \label{fig:mnist_runtime}
\end{subfigure}%
\begin{subfigure}{.6\textwidth}
  \centering
  \includegraphics[width=\linewidth]{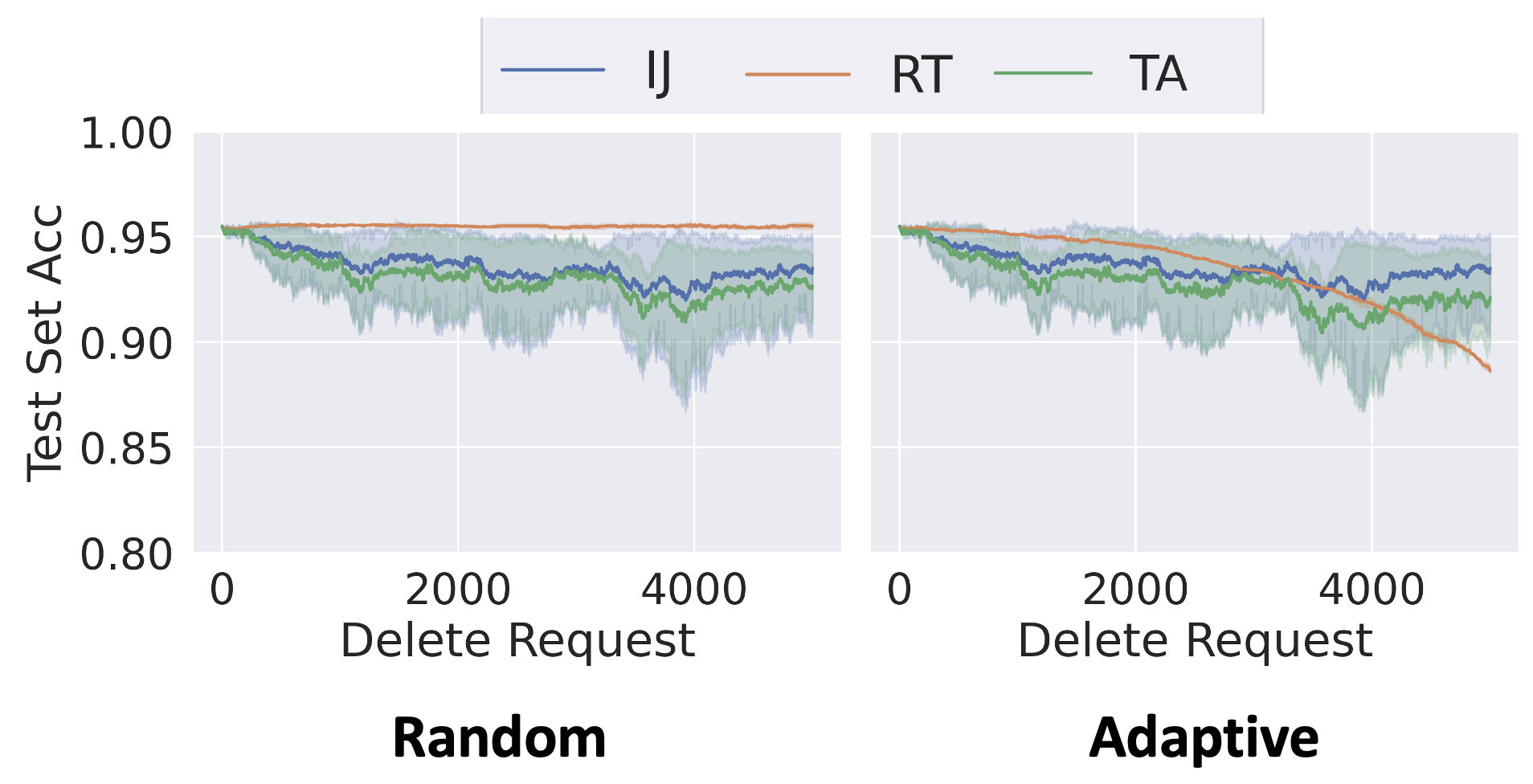}
  \label{fig:mnist_test_acc}
\end{subfigure}
\caption{\textbf{IJ} vs. \textbf{RT} \& \textbf{TA} for smooth regularizers. Comparing both the runtime and test accuracy of the unlearned models in our $\ell_2$ logistic regression setup for predicting 3's and 8's on MNIST. For GDPR, the y-axis of the runtime graph denotes the runtime at month X under the current one month delete request window.}
\label{fig:mnist}
\end{figure}

\paragraph{Logistic Regression with Smooth Regularizers}

In this experiment we simulate an online data deletion setup using the MNIST dataset. For simplicity, we restrict the problem to binary classification by predicting between 3s and 8s. This is the same setup used by~\citet{guo2020certified} to evaluate their approximate unlearning algorithm. %The three data deletion algorithms considered in this experiment are: \textbf{RT}, \textbf{TA}, and our algorithm \textbf{IJ}. 
We flatten the MNIST image into a 1-D vector and train an $\ell_2$ regularized logistic regression model for each algorithm. We evaluate the impact of a deleting a sequence of 5000 datapoints (approximately 40\% of the total dataset) both randomly and in an adaptively chosen manner. In these experiments we consider noise at $c=0.01$.

  On average, \textbf{IJ} was 2611x faster than \textbf{RT} and 2045x faster than \textbf{TA} (Figure~\ref{fig:mnist}). We find that this improvement in runtime comes at very minimal cost to the test performance of the model returned by our algorithm compared to the test performance of \textbf{TA} (Figure~\ref{fig:mnist}). Furthermore, we observe that these results also hold when the delete requests are chosen adaptively (Figure~\ref{fig:mnist}). As discussed previously, the main savings in computation for our algorithm comes from only inverting the hessian once while \textbf{TA} requires a hessian inversion for every delete request.

% \begin{figure}[t]
% \centering
% \includegraphics[width=\textwidth]{figures/final_smooth.png}
% \caption{\textbf{IJ} vs. \textbf{TA} and \textbf{RT}. Comparing both the runtime and test accuracy of the unlearned models in our $\ell_2$ logistic regression setup for predicting 3 vs. 8 on MNIST. \textbf{IJ} significantly improves upon the runtime (C) of both baselines while maintaining similar or better generalization (A and B) compared to \textbf{TA}. }
% \label{fig:smooth}
% \end{figure}

\begin{figure}[h]
\centering
\begin{subfigure}{.25\textwidth}
  \centering
  \includegraphics[width=\linewidth]{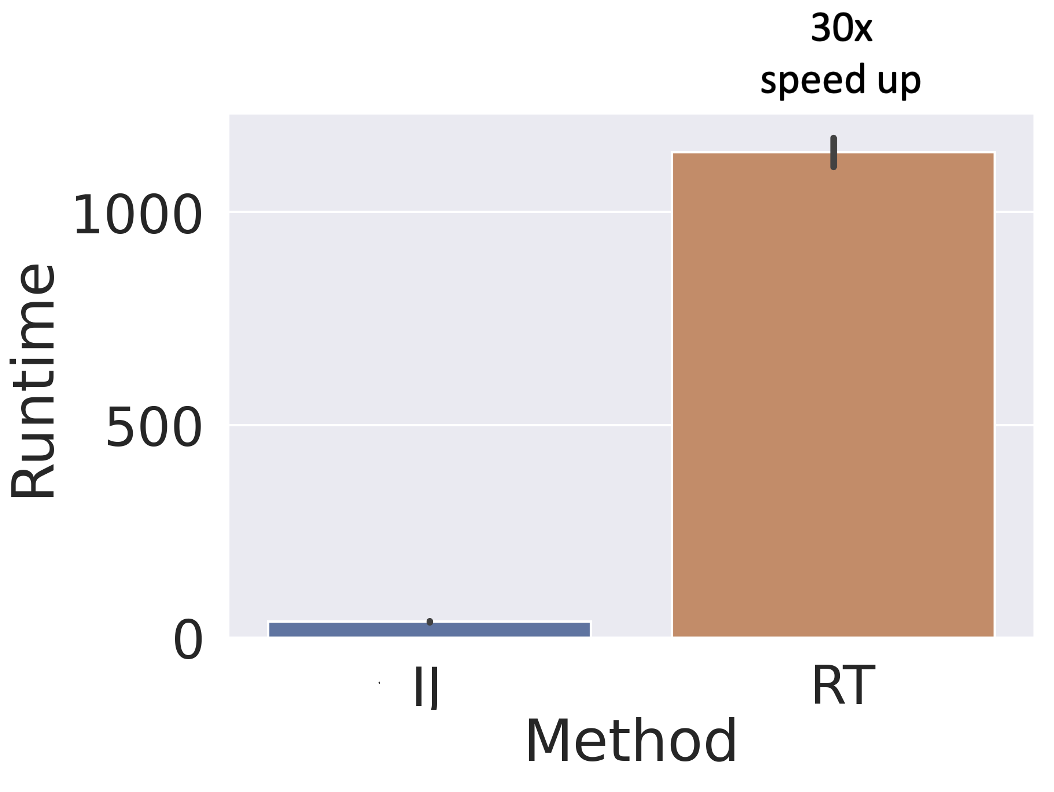}
  \label{fig:prox_runtime}
\end{subfigure}%
\quad\quad\hspace{10pt}
\begin{subfigure}{.25\textwidth}
  \centering
  \includegraphics[width=\linewidth]{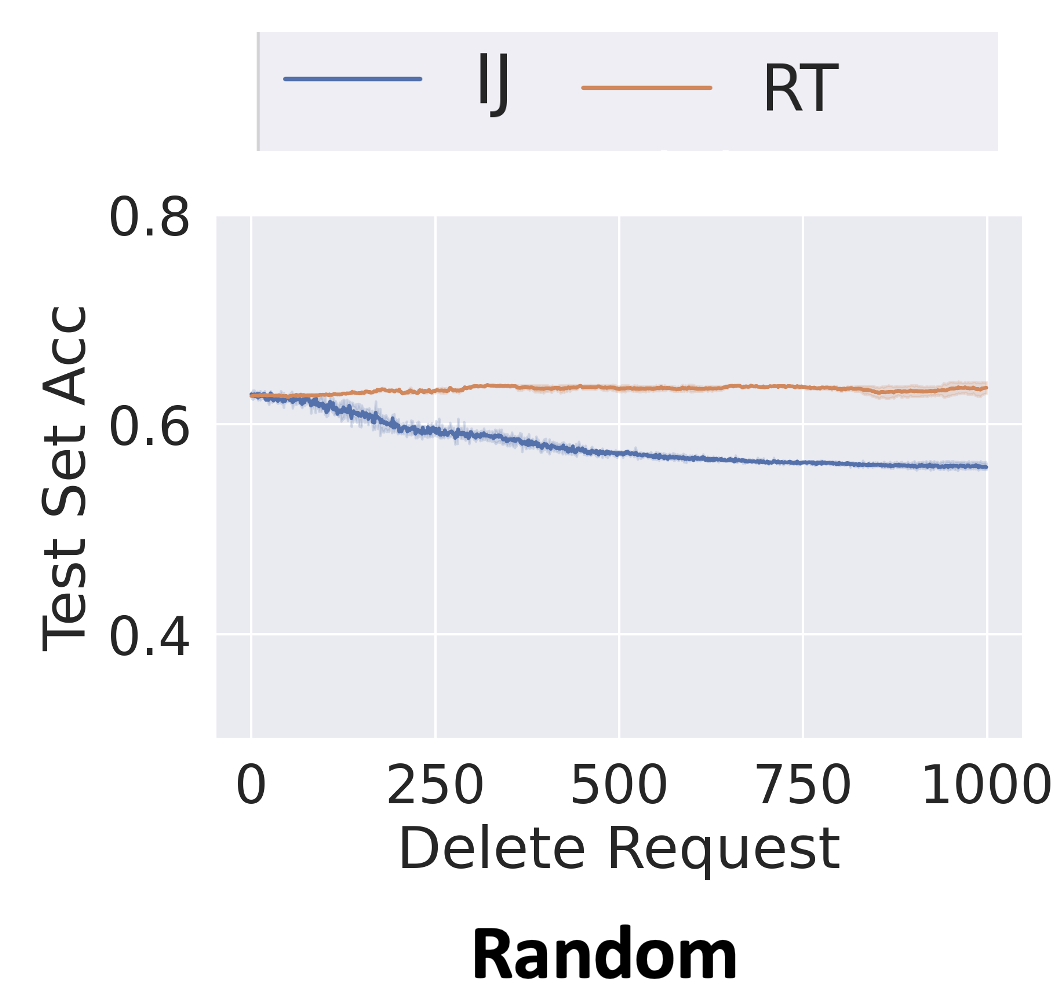}
  \label{fig:prox_test_acc}
\end{subfigure}
\caption{\textbf{IJ} vs. \textbf{RT} for non-smooth regularizers. Comparing both the runtime and test accuracy of the unlearned models in our $\ell_1$ logistic regression setup for predicting warfarin dosage. For GDPR, the y-axis of the runtime graph denotes the runtime at month X under the current one month delete request window.}
\label{fig:prox}
\vspace{-0.7mm}
\end{figure}

\paragraph{Logistic Regression with Non-Smooth Regularizers}

This experiment showcases the performance of our proximal Newton algorithm on a problem with a non-smooth regularizer. We focus on predicting warfarin dosing from patient demographic and physiological data because it a practical setting where $\ell_1$ regularization has been demonstrated to be preferred~\cite{sharabiani2015revisiting}. Given that \textbf{TA} does not naturally support non-smooth regularizers we only compare \textbf{IJ} to \textbf{RT}. In these experiments we consider a logistic regression objective, $\ell_1$ regularizer, and noise level of $c=0.01$.
 On average, \textbf{IJ} was 30x faster than \textbf{RT} (Figure~\ref{fig:prox}). This improvement in runtime only comes at small cost to the test accuracy of the model returned by our algorithm compared to the test performance of \textbf{TA} (Figure~\ref{fig:prox}). %This is similar to that seen in our results with $\ell_2$ regularized logistic regression.
 
 \paragraph{Non-Convex: Logistic Regression with Differentially Private Feature Extractor}
 \begin{figure}[h]
\centering
\begin{subfigure}{.3\textwidth}
  \centering
  \includegraphics[width=\linewidth]{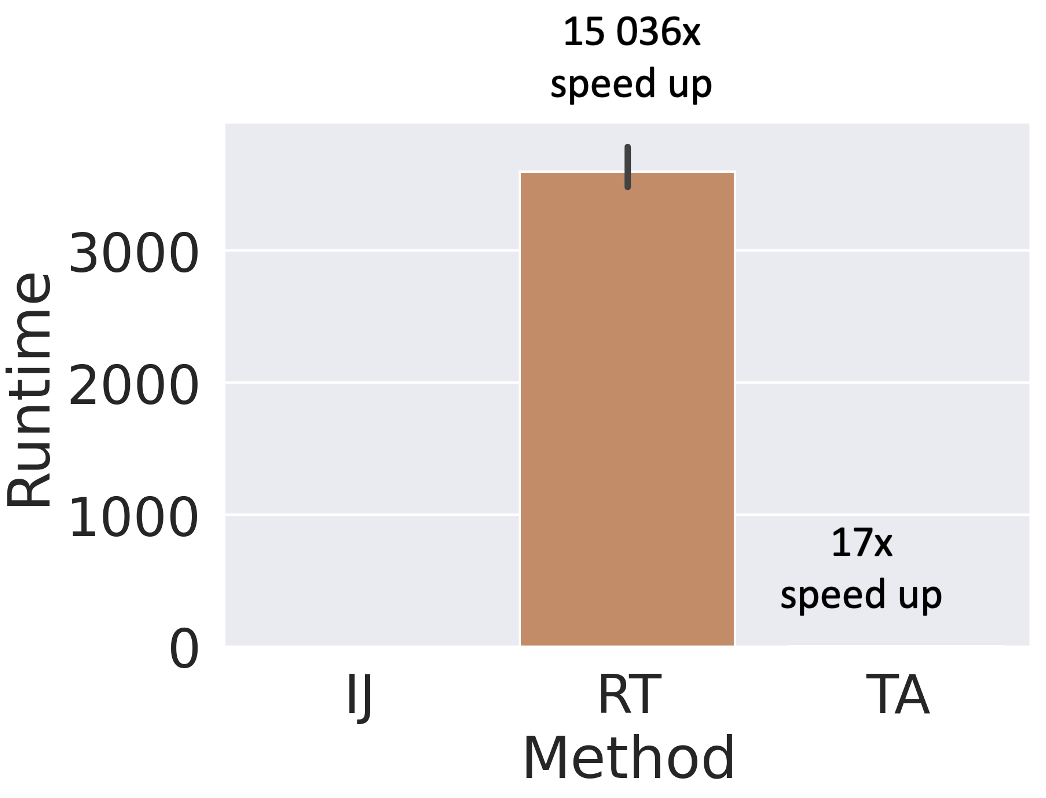}
  \label{fig:sub1}
\end{subfigure}%
\begin{subfigure}{.5\textwidth}
  \centering
  \includegraphics[width=\linewidth]{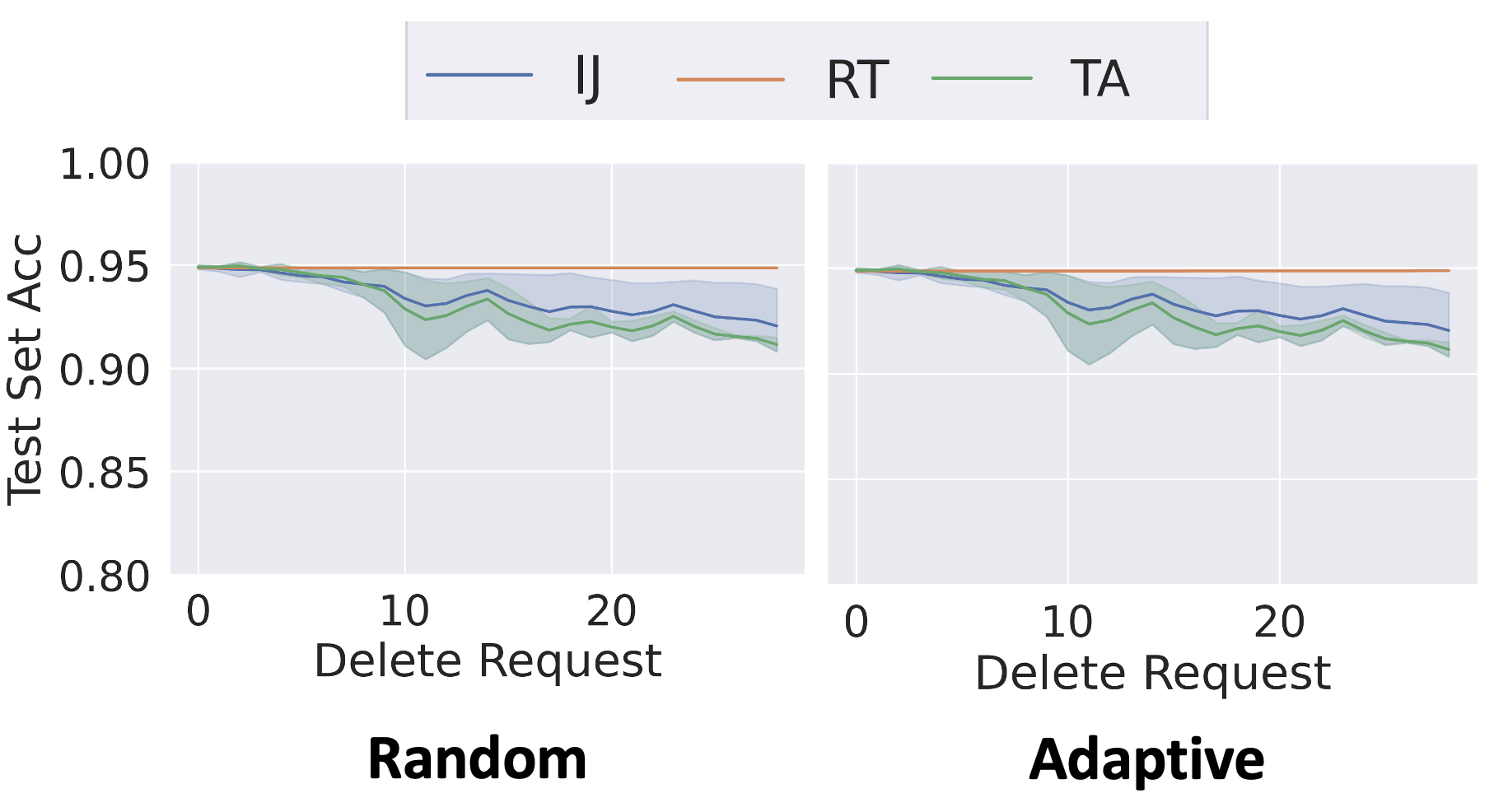}
  \label{fig:sub2}
\end{subfigure}
\caption{\textbf{IJ} vs. \textbf{TA} and \textbf{RT} for non-convex training. Comparing both the runtime and test accuracy of the unlearned models in our DP feature extractor + $\ell_2$ logistic regression setup for predicting digits in photos of street signs from SVHN. For GDPR, the y-axis of the runtime graph denotes the runtime at month X under the current one month delete request window.}
\label{fig:svhn}
\vspace{-0.5mm}
\end{figure}
 We demonstrate the ability to use our unlearning algorithm in non-convex settings. Similar to~\citet{guo2020certified} we train a differentially private feature extractor with $\epsilon = 0.1$ on street digit signs from SVHN to extract representations which a logistic regression model can be applied on top of.  Similar to previous experiments, we pick digits 3 and 8 for simplicity and  observe significant speedups using \textbf{IJ}. On average, \textbf{IJ} was 15036x faster than \textbf{RT} and 17x faster than \textbf{TA} and these speedups come at a marginal cost to the test accuracy of our algorithm even as the number of delete requests increase (Figure~\ref{fig:svhn}).

\section{Unlearning hyperparameter-tuned models}
\label{sec:failure_mode}
One of the most common techniques for hyperparameter tuning is {\em cross-validation}. Here datapoints are used to validate and select models using the following cross-validation error objective
\begin{talign}\label{eq:cverror}
\text{CV}(\lambda) = \frac{1}{n}\sum_{i=1}^n \ell(z_i, \hat{\theta}_{n,-i}(\lambda)).
\end{talign}
Specifically, the model selection pipeline oftentimes entails selecting $\lambda \in \Lambda$ that minimizes the CV error~\eqref{eq:cverror}. Given each datapoint is used to select models, one might hope  it is possible to unlearn models using~\cref{algo:1} when CV is used to select the model. However, as the following proposition illustrates, the $(\epsilon,\delta)$-unlearning guarantees of Theorem~\ref{thm:1} no longer apply when CV has taken place. %this kind of data-driven hyperparameter tuning. 

\begin{proposition}\label{prop:1} Suppose cross-validation is used to select $\lambda \in \Lambda$ and the empirical risk minimizer $\estt$ is returned. Consider a delete request by user $i$ and the model returned by unlearning procedure~\eqref{eq:s} or~\eqref{algo:1}, which we denote $\tilde{\theta}_{n,-i}(\lambda)$. Then it is possible $\|\tilde{\theta}_{n,-i}(\lambda) - \hat{\theta}_{n,-i}(\lambda')\| = o(1/n)$ where $\hat{\theta}_{n,-i}(\lambda')$ is the model selected after deleting user $i$ and performing cross-validation again to select $\lambda' \in \Lambda$ before returning the empricial risk minizer. 

\end{proposition}
{\em Proof of~\cref{prop:1}.}  
Suppose $\ell(z,\theta) = \frac{1}{2}(z - \theta)^2$, $\pi(\theta) = \theta^2$ and $\Lambda = \{0, \infty\}$. Note that $\hat{\theta}_n(\lambda) = \frac{1}{1+\lambda} \bar{z}$ where $\bar{z} = \frac{1}{n} \sum_{i=1}^n z_i$ is the sample average. Consider a dataset consisting of $n-1$ points (group A) with value $-\frac{1}{n}$ and a remaining point (group B) with value $n$ with $n\geq2$, and suppose the user in group B requests to delete their data. 
%
%We will show cross-validation will result in $\lambda = \infty$ and the %and sample average $\hat{\theta}_{n}(\infty) = 0$. A delete requiest from the user in group $B$ will result in 
%leave-one-out estimator $\tilde{\theta}_{n,-i}(\infty) = 0$, but cross-validation performed with the dataset sans the datapoint from group B would result in $\lambda = 0$ and empirical risk minimizer $\hat{\theta}_{n,-i}(0) = -\frac{1}{n}$. That is, we will show that in this example,  $\|\tilde{\theta}_{n,-i}(\infty)- \hat{\theta}_{n,-i}(0) \| = o(1/n)$ providing a counter-example.
Performing cross-validation on this dataset will entail computing two kinds of sample averages, one where we have deleted a point from group $A$, $\bar{z}_{-i_A} = \frac{1}{n-1}(-\frac{n-2}{n} + n)$, and one where we have deleted the point from group $B$, $\bar{z}_{-i_B} =-\frac{1}{n}$. Recall the CV error 
\begin{talign*}
    \text{CV}(\lambda) = \frac{1}{2n} \sum_{i=1}^n (z_i - \hat{\theta}_{n,-i}(\lambda))^2 = \frac{n-1}{2n}(\frac{1}{1+\lambda}\frac{1}{n-1}(n-\frac{n-2}{n} )-\frac{1}{n})^2 + \frac{1}{2n}(n + \frac{1}{1+\lambda}\frac{1}{n})^2
\end{talign*}
When minimized over the set $\Lambda = \{0,\infty\}$, $\lambda = \infty$ is optimal for any $n\geq2$. This results in estimator $\hat{\theta}_{n}(\lambda) =\hat{\theta}_{n}(\infty) = 0$. The leave-one-out approximation is the same for estimators~\eqref{algo:1} and~\eqref{eq:s}, given $\nabla_{\theta}^2 F(z,\theta,\lambda)^{-1} = \frac{1}{1 + \lambda}$. Subsequently, $\tilde{\theta}_{n,-i}(\lambda) = \hat{\theta}_{n}(\lambda) + \frac{1}{n}\frac{1}{1+\lambda}( \hat{\theta}_{n}(\lambda) - z_i )$ results in the approximation $\tilde{\theta}_{n,-i}(\infty) = 0$. On the other hand, having deleted the datapoint from group $B$, we have a new sample mean $\bar{z} = -\frac{1}{n}$ and CV error, given by 
\begin{talign*}
\text{CV}(\lambda) = \frac{1}{2n} \sum_{i=1}^n (z_i - \hat{\theta}_{n,-i}(\lambda))^2 = \frac{1}{2n}(-\frac{1}{n} + \frac{1}{1+\lambda} \frac{1}{n})^2
\end{talign*}
is minimized by $\lambda = 0$. This results in leave-one-out minimizer $\hat{\theta}_{n,-i}(0) = -\frac{1}{n}$. $\qedsymbol$

%\begin{corollary}
An implication of~\cref{prop:1} is that~\cref{algo:1} as well as the algorithms proposed by~\cite{sekhari2021remember,guo2020certified}  do not unlearn models when cross-validation is used to selected hyperparameters.  This is because the noise term $\sigma \propto O(m^2/n^2)$ is no longer sufficient to guarantee datapoint $i$ is unlearned. %if the variance of the Gaussian noise scales as $c = o(1/n^2)$}. 
This means that organizations which train their models using cross-validation and use approximate unlearning algorithms could still be leaking information about the data which they delete.
%\end{corollary}
%One might wonder whether this is tight. 

\section{Discussion}

%\paragraph{Importance of Online Data Deletion Algorithms} 
A main contribution of our study is the development of an efficient online/batch data deletion algorithm for non-smooth problems. The only previous work proposing a low-memory online algorithm is~\citet{guo2020certified}. However,~\citet{marchant2021hard} show that this algorithm is susceptible to poisoning attacks rendering it highly inefficient. %provide an online algorithm it suffers from the potential to be as inefficient as retraining (evidenced by the poisoning attacks of~\cite{marchant2021hard}).
%Given GDPR will mean that organizations are going to (1) constantly be receiving delete requests from users throughout time and (2) will be required to continually satisfy request, an efficient online algorithm could be useful. %. This setup naturally requires online algorithms as it will almost never be the case where an organization receives all delete requests at once. 
While exact unlearning algorithms such as SISA and retraining naturally work in an online setting, our work is the first to make this work for approximate unlearning algorithms. Further development of data deletion algorithms should focus more minimizing computational cost on the streaming batch setting to ensure practical use. Also, while the current form of GDPR legislation allows quite a bit of time for companies to comply (30 days), we believe that showing a tool can provide the same empirical performance and theoretical guarantees with immediate deletion is beneficial toward encouraging companies to complying with GDPR requests more swiftly (and might encourage lawmakers to necessitate faster compliance). When individuals request to delete their data, it is sometimes because they are concerned about the risk of potential harm if their data remains available; this risk is potentially compounded the longer it takes for the data to be deleted. By showing it can be done quickly, companies may be encouraged to act more expeditiously and less harm might occur.

% \paragraph{cross-validation and Data Deletion} 
The infinitesimal jacknife has been previously been used to perform cross-validation. We note deep connections between the approximate cross-validation algorithms and approximate machine unlearning algorithms. The algorithm developed by~\cite{sekhari2021remember}, for example, can be viewed as an analog of the approximate cross-validation algorithm proposed by~\cite{beirami2017optimal, rad2019scalable}. Additionally, the model selection error bounds provided in the approximate cross-validation literature for example in~\cite{wilson2020approximate} are very similar to the generalization guarantees proved in our work and~\cite{sekhari2021remember} where we are concerned with the error introduced by our approximation to the unlearning baseline. %In approximate cross-validation this is synonymous with the error between true cross-validation and the approximate cross-validation algorithm proposed. 
 We believe further connections could be exploited to provide unlearning algorithms when hyperparameter tuning has taken place. %We note that we are the first study to our knowledge to explicitly draw this important connection and we encourage researchers working on cross-validation (and its approximate forms) to work together with those working on data deletion to help improve results in both fields.

%\paragraph{Parametric Definitions of Data Deletion} 
 In this work and in several others we consider a definition of data deletion that is parameterized by two values $\epsilon$ and $\delta$. As is the case with differential privacy, it is currently unclear what the impact of satisfying different levels of $\epsilon$ means practically. Developing auditing algorithms~\cite{thudi2021necessity} similar to those recently seen in the DP community~\cite{NEURIPS2020_fc4ddc15, nasr2021adversary} can help provide more transparency on the meaning of the $(\epsilon, \delta)$-unlearning guarantee.

%\paragraph{What Does it Mean to Delete Data?} 
Finally, we return to the broader question we posed at the beginning: \textit{what does it mean to delete data from a  machine learning pipeline?} Most existing work focuses on data deletion in the model training process. Yet, the machine learning and data analysis pipeline is much broader than just model training. As evidenced in Section~\ref{sec:failure_mode}, the lack of research on data deletion in model selection could potentially result in information being leaked from previously proposed unlearning algorithms. Given this result, it is likely that all unlearning algorithms ({\em both exact and inexact}) still leak information about deleted data in other stages of the machine learning pipeline such as exploratory data analysis and feature selection. We encourage the community to explore definitions of data deletion which encompass the entire machine learning pipeline.

\section{Acknowledgements} The authors would like to thank the NeurIPS 2022 reviewers and area chair for their comments and feedback. The authors acknowledge the MIT SuperCloud and Lincoln Laboratory Supercomputing Center for providing (HPC, database, consultation) resources that have contributed to the research results reported within this paper/report. VMS is supported by a Wellcome Trust Fellowship.

\clearpage
\bibliography{refs}

\begin{thebibliography}{35}
\providecommand{\natexlab}[1]{#1}
\providecommand{\url}[1]{\texttt{#1}}
\expandafter\ifx\csname urlstyle\endcsname\relax
  \providecommand{\doi}[1]{doi: #1}\else
  \providecommand{\doi}{doi: \begingroup \urlstyle{rm}\Url}\fi

\bibitem[Abadi et~al.(2016)Abadi, Chu, Goodfellow, McMahan, Mironov, Talwar,
  and Zhang]{abadi2016deep}
Martin Abadi, Andy Chu, Ian Goodfellow, H~Brendan McMahan, Ilya Mironov, Kunal
  Talwar, and Li~Zhang.
\newblock Deep learning with differential privacy.
\newblock In \emph{Proceedings of the 2016 ACM SIGSAC conference on computer
  and communications security}, pages 308--318, 2016.

\bibitem[Bassily et~al.(2019)Bassily, Feldman, Talwar, and
  Guha~Thakurta]{bassily2019private}
Raef Bassily, Vitaly Feldman, Kunal Talwar, and Abhradeep Guha~Thakurta.
\newblock Private stochastic convex optimization with optimal rates.
\newblock \emph{Advances in Neural Information Processing Systems}, 32, 2019.

\bibitem[Beirami et~al.(2017)Beirami, Razaviyayn, Shahrampour, and
  Tarokh]{beirami2017optimal}
Ahmad Beirami, Meisam Razaviyayn, Shahin Shahrampour, and Vahid Tarokh.
\newblock On optimal generalizability in parametric learning.
\newblock In \emph{Advances in Neural Information Processing Systems}, pages
  3455--3465, 2017.

\bibitem[Bourtoule et~al.(2021)Bourtoule, Chandrasekaran, Choquette-Choo, Jia,
  Travers, Zhang, Lie, and Papernot]{bourtoule2021machine}
Lucas Bourtoule, Varun Chandrasekaran, Christopher~A Choquette-Choo, Hengrui
  Jia, Adelin Travers, Baiwu Zhang, David Lie, and Nicolas Papernot.
\newblock Machine unlearning.
\newblock In \emph{2021 IEEE Symposium on Security and Privacy (SP)}, pages
  141--159. IEEE, 2021.

\bibitem[Bygrave(2014)]{bygrave2014right}
Lee~A Bygrave.
\newblock A right to be forgotten?
\newblock \emph{Communications of the ACM}, 58\penalty0 (1):\penalty0 35--37,
  2014.

\bibitem[Cao and Yang(2015)]{cao2015towards}
Yinzhi Cao and Junfeng Yang.
\newblock Towards making systems forget with machine unlearning.
\newblock In \emph{2015 IEEE Symposium on Security and Privacy}, pages
  463--480. IEEE, 2015.

\bibitem[Commission()]{FTC}
Federal~Trade Commission.
\newblock California company settles ftc allegations it deceived consumers
  about use of facial recognition in photo storage app.

\bibitem[Consortium(2009)]{international2009estimation}
International Warfarin~Pharmacogenetics Consortium.
\newblock Estimation of the warfarin dose with clinical and pharmacogenetic
  data.
\newblock \emph{New England Journal of Medicine}, 360\penalty0 (8):\penalty0
  753--764, 2009.

\bibitem[Dwork et~al.(2014)Dwork, Roth, et~al.]{dwork2014algorithmic}
Cynthia Dwork, Aaron Roth, et~al.
\newblock The algorithmic foundations of differential privacy.
\newblock \emph{Found. Trends Theor. Comput. Sci.}, 9\penalty0 (3-4):\penalty0
  211--407, 2014.

\bibitem[Efron(1982)]{efron1982jackknife}
Bradley Efron.
\newblock \emph{The jackknife, the bootstrap and other resampling plans}.
\newblock SIAM, 1982.

\bibitem[Garg et~al.(2020)Garg, Goldwasser, and Vasudevan]{garg2020formalizing}
Sanjam Garg, Shafi Goldwasser, and Prashant~Nalini Vasudevan.
\newblock Formalizing data deletion in the context of the right to be
  forgotten.
\newblock In \emph{Annual International Conference on the Theory and
  Applications of Cryptographic Techniques}, pages 373--402. Springer, 2020.

\bibitem[Ginart et~al.(2019)Ginart, Guan, Valiant, and Zou]{ginart2019making}
Antonio Ginart, Melody Guan, Gregory Valiant, and James~Y Zou.
\newblock Making ai forget you: Data deletion in machine learning.
\newblock \emph{Advances in Neural Information Processing Systems}, 32, 2019.

\bibitem[Giordano et~al.(2019)Giordano, Stephenson, Liu, Jordan, and
  Broderick]{giordano2019swiss}
Ryan Giordano, William Stephenson, Runjing Liu, Michael Jordan, and Tamara
  Broderick.
\newblock A swiss army infinitesimal jackknife.
\newblock In \emph{The 22nd International Conference on Artificial Intelligence
  and Statistics}, pages 1139--1147. PMLR, 2019.

\bibitem[Guo et~al.(2020)Guo, Goldstein, Hannun, and Van
  Der~Maaten]{guo2020certified}
Chuan Guo, Tom Goldstein, Awni Hannun, and Laurens Van Der~Maaten.
\newblock Certified data removal from machine learning models.
\newblock In \emph{International Conference on Machine Learning}, pages
  3832--3842. PMLR, 2020.

\bibitem[Gupta et~al.(2021)Gupta, Jung, Neel, Roth, Sharifi-Malvajerdi, and
  Waites]{gupta2021adaptive}
Varun Gupta, Christopher Jung, Seth Neel, Aaron Roth, Saeed Sharifi-Malvajerdi,
  and Chris Waites.
\newblock Adaptive machine unlearning.
\newblock \emph{Advances in Neural Information Processing Systems}, 34, 2021.

\bibitem[Izzo et~al.(2021)Izzo, Smart, Chaudhuri, and Zou]{izzo2021approximate}
Zachary Izzo, Mary~Anne Smart, Kamalika Chaudhuri, and James Zou.
\newblock Approximate data deletion from machine learning models.
\newblock In \emph{International Conference on Artificial Intelligence and
  Statistics}, pages 2008--2016. PMLR, 2021.

\bibitem[Jaeckel(1972)]{jaeckel1972infinitesimal}
L~Jaeckel.
\newblock The infinitesimal jackknife. memorandum.
\newblock Technical report, MM 72-1215-11, Bell Lab. Murray Hill, NJ, 1972.

\bibitem[Jagielski et~al.(2020)Jagielski, Ullman, and
  Oprea]{NEURIPS2020_fc4ddc15}
Matthew Jagielski, Jonathan Ullman, and Alina Oprea.
\newblock Auditing differentially private machine learning: How private is
  private sgd?
\newblock In H.~Larochelle, M.~Ranzato, R.~Hadsell, M.F. Balcan, and H.~Lin,
  editors, \emph{Advances in Neural Information Processing Systems}, volume~33,
  pages 22205--22216. Curran Associates, Inc., 2020.
\newblock URL
  \url{https://proceedings.neurips.cc/paper/2020/file/fc4ddc15f9f4b4b06ef7844d6bb53abf-Paper.pdf}.

\bibitem[LeCun(1998)]{lecun1998mnist}
Yann LeCun.
\newblock The mnist database of handwritten digits.
\newblock \emph{http://yann. lecun. com/exdb/mnist/}, 1998.

\bibitem[Marchant et~al.(2021)Marchant, Rubinstein, and
  Alfeld]{marchant2021hard}
Neil~G Marchant, Benjamin~IP Rubinstein, and Scott Alfeld.
\newblock Hard to forget: Poisoning attacks on certified machine unlearning.
\newblock \emph{arXiv preprint arXiv:2109.08266}, 2021.

\bibitem[McGoldrick(2013)]{mcgoldrick2013developments}
Dominic McGoldrick.
\newblock Developments in the right to be forgotten.
\newblock \emph{Human Rights Law Review}, 13\penalty0 (4):\penalty0 761--776,
  2013.

\bibitem[Nasr et~al.(2021)Nasr, Songi, Thakurta, Papemoti, and
  Carlin]{nasr2021adversary}
Milad Nasr, Shuang Songi, Abhradeep Thakurta, Nicolas Papemoti, and Nicholas
  Carlin.
\newblock Adversary instantiation: Lower bounds for differentially private
  machine learning.
\newblock In \emph{2021 IEEE Symposium on Security and Privacy (SP)}, pages
  866--882. IEEE, 2021.

\bibitem[Neel et~al.(2021)Neel, Roth, and Sharifi-Malvajerdi]{neel2021descent}
Seth Neel, Aaron Roth, and Saeed Sharifi-Malvajerdi.
\newblock Descent-to-delete: Gradient-based methods for machine unlearning.
\newblock In \emph{Algorithmic Learning Theory}, pages 931--962. PMLR, 2021.

\bibitem[Netzer et~al.(2011)Netzer, Wang, Coates, Bissacco, Wu, and
  Ng]{netzer2011reading}
Yuval Netzer, Tao Wang, Adam Coates, Alessandro Bissacco, Bo~Wu, and Andrew~Y
  Ng.
\newblock Reading digits in natural images with unsupervised feature learning.
\newblock 2011.

\bibitem[Papernot and Steinke(2021)]{papernot2021hyperparameter}
Nicolas Papernot and Thomas Steinke.
\newblock Hyperparameter tuning with renyi differential privacy.
\newblock \emph{arXiv preprint arXiv:2110.03620}, 2021.

\bibitem[Rad and Maleki(2019)]{rad2019scalable}
Kamiar~Rahnama Rad and Arian Maleki.
\newblock A scalable estimate of the out-of-sample prediction error via
  approximate leave-one-out.
\newblock \emph{arXiv preprint arXiv:1801.10243}, 2019.

\bibitem[Rosen(2011)]{rosen2011right}
Jeffrey Rosen.
\newblock The right to be forgotten.
\newblock \emph{Stan. L. Rev. Online}, 64:\penalty0 88, 2011.

\bibitem[Sekhari et~al.(2021)Sekhari, Acharya, Kamath, and
  Suresh]{sekhari2021remember}
Ayush Sekhari, Jayadev Acharya, Gautam Kamath, and Ananda~Theertha Suresh.
\newblock Remember what you want to forget: Algorithms for machine unlearning.
\newblock \emph{Advances in Neural Information Processing Systems}, 34, 2021.

\bibitem[Shalev-Shwartz et~al.(2009)Shalev-Shwartz, Shamir, Srebro, and
  Sridharan]{shalev2009stochastic}
Shai Shalev-Shwartz, Ohad Shamir, Nathan Srebro, and Karthik Sridharan.
\newblock Stochastic convex optimization.
\newblock In \emph{COLT}, volume~2, page~5, 2009.

\bibitem[Sharabiani et~al.(2015)Sharabiani, Bress, Douzali, and
  Darabi]{sharabiani2015revisiting}
Ashkan Sharabiani, Adam Bress, Elnaz Douzali, and Houshang Darabi.
\newblock Revisiting warfarin dosing using machine learning techniques.
\newblock \emph{Computational and mathematical methods in medicine}, 2015,
  2015.

\bibitem[Shokri et~al.(2017)Shokri, Stronati, Song, and
  Shmatikov]{shokri2017membership}
Reza Shokri, Marco Stronati, Congzheng Song, and Vitaly Shmatikov.
\newblock Membership inference attacks against machine learning models.
\newblock In \emph{2017 IEEE symposium on security and privacy (SP)}, pages
  3--18. IEEE, 2017.

\bibitem[Thudi et~al.(2021)Thudi, Jia, Shumailov, and
  Papernot]{thudi2021necessity}
Anvith Thudi, Hengrui Jia, Ilia Shumailov, and Nicolas Papernot.
\newblock On the necessity of auditable algorithmic definitions for machine
  unlearning.
\newblock \emph{arXiv preprint arXiv:2110.11891}, 2021.

\bibitem[Ullah et~al.(2021)Ullah, Mai, Rao, Rossi, and Arora]{ullah2021machine}
Enayat Ullah, Tung Mai, Anup Rao, Ryan~A Rossi, and Raman Arora.
\newblock Machine unlearning via algorithmic stability.
\newblock In \emph{Conference on Learning Theory}, pages 4126--4142. PMLR,
  2021.

\bibitem[Voigt and Von~dem Bussche(2017)]{voigt2017eu}
Paul Voigt and Axel Von~dem Bussche.
\newblock The eu general data protection regulation (gdpr).
\newblock \emph{A Practical Guide, 1st Ed., Cham: Springer International
  Publishing}, 10\penalty0 (3152676):\penalty0 10--5555, 2017.

\bibitem[Wilson et~al.(2020)Wilson, Kasy, and Mackey]{wilson2020approximate}
Ashia Wilson, Maximilian Kasy, and Lester Mackey.
\newblock Approximate cross-validation: Guarantees for model assessment and
  selection.
\newblock In \emph{International Conference on Artificial Intelligence and
  Statistics}, pages 4530--4540. PMLR, 2020.

\end{thebibliography}

\newpage
\appendix

\section{Proof of (\ref{thm:1})}
\begin{lemma}[Optimization comparison lemma~\cite{wilson2020approximate}]
\label{lem:optcomp}
Suppose
\begin{align}\label{eq:obj} 
x^\ast \in \argmin_x \varphi_1(x) + \varphi_0(x) \quad \text{and} \quad  y^\ast \in \argmin_x \varphi_2(x) + \varphi_0(x).
\end{align}
for $\varphi_1$ and $\varphi_2$ differentiable and $\varphi_0$ convex.
\end{lemma}
\begin{proof}
The (sub)differentiability assumptions and the optimality of $x_{\varphi_1}$ and $x_{\varphi_2}$ imply that $0 \in \partial \varphi_2$ and $u = 0 + \nabla(\varphi_1 - \varphi_2)(x_{\varphi_1})$ for some $ u \in x_{\varphi_2}$. The gradient growth condition implies 
\balignt
\nu_{\varphi_2}(\twonorm{x_{\varphi_1} - x_{\varphi_2} })
    \leq 
    \langle x_{\varphi_1} - x_{\varphi_2},u - 0\rangle
    = 
    \langle x_{\varphi_1} - x_{\varphi_2},\nabla (\varphi_2- \varphi_1)(x_{\varphi_1})\rangle.
\ealignt
\end{proof}
\begin{lemma}[Learning guarantee for $\hat{\theta}_{n}(\lambda)$]\label{lemma:shalev}
Given, $F_{n}$ satisfies Assumption~\ref{ass:1} or \ref{ass:2} and any distribution $\mathcal{D}$, let $S=\{z_{i}\}_{i=1}^{n}$ where $S \sim \mathcal{D}^{n}$. Then the the empirical minimizer $\hat{\theta}_{n}(\lambda)$ of $F_n(\theta, \lambda, z)$ satisfies

$$\mathbb{E}[F(\hat{\theta}_{n}(\lambda)) - F(\theta^{*}(\lambda))] \leq \frac{4L^2}{\mu n}$$

\begin{proof}
Given $F_{n}$ is $\mu$-strongly convex this follows from Claim 6.2 in~\cite{shalev2009stochastic}.
\end{proof}

\end{lemma}

% First, we prove the learning guarantee of any empirical risk minimizer given Assumption~\ref{ass:1}.

% \begin{lemma}[Learning guarantee for $\hat{\theta_{n}}$]
% For any distribution $\mathcal{D}$, :

% $$\mathbb{E}[F(\hat{\theta_n })] - F(\theta^{*}) \leq \frac{4L^2}{\lambda n}$$

% \begin{proof}
% Since $F$ is $L$-Lipschitz and $\lambda$-strongly convex the guarantee follows from Lemma~\ref{lemma:shalev}.
% \end{proof}

\subsection{Proof of~\eqref{eq:ineq1}: Closeness of $\hat{\theta}_n(\lambda)$ and $\hat{\theta}_{n, -U}(\lambda)$}
\label{app:closeness_proof}
Suppose we have deleted $m$ users in a set $U$. Define $\tilde{F}_{n,-U} = \frac{n-m}{n} F_{n,-U}$ where $F_{n,-U} = \frac{1}{n-m}\sum_{i \not \in U} f(z_i, \theta, \lambda)$ and note that $\tilde{F}_{n,-U}$ and $ F_{n,-U} $ have the same minimizers. We will work with $\tilde{F}_{n,-U} $. 
By the optimizer comparison lemma~\ref{lem:optcomp} and strong convexity of $F_{n}$
\begin{talign*}
\mu\twonorm{ \hat{\theta}_n(\lambda)- \hat{\theta}_{n, -U}(\lambda)}^2 &\leq \langle \hat{\theta}_n(\lambda)- \hat{\theta}_{n, -U}(\lambda), \nabla F_{n}(z,\hat{\theta}_n(\lambda), \lambda)  - \nabla \tilde{F}_{n,-U}(z, \hat{\theta}_n(\lambda), \lambda)\rangle\\
& = \frac{1}{n}\sum_{i\in U}\langle \hat{\theta}_n(\lambda)- \hat{\theta}_{n, -U}(\lambda), \nabla \ell(z_i,\hat{\theta}_n(\lambda))\rangle \\
&\leq \frac{1}{n}\|\hat{\theta}_n(\lambda)- \hat{\theta}_{n, -U}(\lambda) \|_2\sum_{i\in U}\|\nabla \ell(z_i,\hat{\theta}_n(\lambda))\|_2\\
&\leq \frac{1}{n}\|\hat{\theta}_n(\lambda)- \hat{\theta}_{n, -U}(\lambda) \|_2 \cdot m L 
\end{talign*}
 Dividing both sides by $\| \hat{\theta}_n(\lambda)- \hat{\theta}_{n, -U}(\lambda)\|_2$ and rearranging gives the desired bound of 
 $$ \twonorm{ \hat{\theta}_n(\lambda)- \hat{\theta}_{n, -U}(\lambda)} 
 \leq \frac{m L}{\mu n}$$ 
 
\subsubsection{Proof of~\eqref{eq:ineq1}: Closeness of $\hat{\theta}_{n,-U}(\lambda)$ and $\bar{\theta}_{n, -U}(\lambda)$}
We define:
\begin{itemize}
    \item $\psi_{1} = \tilde{F}_{n,-U}(z, \theta, \lambda)$
    \item $\psi_{2} = \langle \nabla \tilde{F}_{n, -U}(\estt), \estt - \theta \rangle + \langle \estt - \theta, \nabla_{\theta}^{2}\tilde{F}_{n, -U}(\estt)[\estt - \theta]\rangle$
    \item $\psi_{3} = \langle \nabla \tilde{F}_{n, -U}(\estt), \estt - \theta \rangle + \langle \estt - \theta, \nabla_{\theta}^{2}\tilde{F}_n(\estt)[\estt - \theta]\rangle$
    \item $\hat{\theta}_{n,-U}(\lambda) = \argmin \psi_{1}(\theta)$, 
    \item $\Tilde{\theta}_{n,-U}(\lambda) = \argmin\psi_{3}(\theta)$
\end{itemize}   
The optimizer comparison theorem and strong convexity of $F_n$ implies the following upper bound:
\begin{talign*}
\frac{\mu}{2}\|\hat{\theta}_{n,-U}(\lambda) - \tilde{\theta}_{n,-U}(\lambda) \|_2^{2} &\leq \langle \hat{\theta}_{n,-U}(\lambda) - \tilde{\theta}_{n,-U}(\lambda), \nabla(\psi_{3} - \psi_{1})(\hat{\theta}_{n,-U}(\lambda))\rangle \\
&\leq \|\hat{\theta}_{n,-U}(\lambda) - \tilde{\theta}_{n,-U}(\lambda)\|_2\|\nabla(\psi_{3} - \psi_{1})(\hat{\theta}_{n,-U}(\lambda))\|_2 
\end{talign*}
Dividing both sides by $\|\hat{\theta}_{n,-U}(\lambda) - \tilde{\theta}_{n,-U}(\lambda) \|_2$ gives
\begin{talign*}
%\frac{\mu}{2}\|\hat{\theta}_{n,-U}(\lambda) - \tilde{\theta}_{n,-U}(\lambda)\| &\leq \|\nabla(\psi_{3} - \psi_{1})(\hat{\theta}_{n,-U}(\lambda))\| \\
\frac{\mu}{2}\|\hat{\theta}_{n,-U}(\lambda) - \tilde{\theta}_{n,-U}(\lambda) \|_2 &\leq \| \nabla_\theta(\psi_{3}- \psi_{2})(\hat{\theta}_{n,-U}(\lambda)) -  \nabla_\theta(\psi_{2} - \psi_1)(\hat{\theta}_{n,-U}(\lambda))  \|_2 \\
&\leq \| \nabla_\theta(\psi_{3}- \psi_{2})(\hat{\theta}_{n,-U}(\lambda))\|_2  + \|\nabla_\theta(\psi_{2} - \psi_1)(\hat{\theta}_{n,-U}(\lambda)) \|_2 \\
& \leq \|\nabla_{\theta}^{2}\tilde{F}_{n}(\hat{\theta}_{n,-U}(\lambda)) - \nabla_{\theta}^{2}\tilde{F}_{n,-U}(\hat{\theta}_{n,-U}(\lambda))\|_2\|\estt - \hat{\theta}_{n,-U}(\lambda)\|_2\\
&\quad + \|\nabla_\theta(\psi_{2} - \psi_1)(\hat{\theta}_{n,-U}(\lambda)) \|_2 \\
& \leq \frac{m^2 C L}{\mu n^2} + \|\nabla_\theta \psi_{2}(\hat{\theta}_{n,-U}(\lambda))  - \nabla_\theta\psi_{1}(\hat{\theta}_{n,-U}(\lambda)) \|_2 \\
& \overset{\tiny{\textcircled{1}}}{\leq} \frac{m^2 C L}{\mu n^2} + \frac{M}{2}\|\hat{\theta}_{n,-U}(\lambda) - \estt\|_2^2 \\
& \leq \frac{m^2 C L}{\mu n^2} + \frac{M}{2}\cdot \frac{m^2 L^2}{\mu^2 n^2} 
%& \leq \frac{m^2 C L}{\mu n^2} + \frac{Mm^2 L^2}{2 \mu^2 n^2}
\end{talign*}
Inequality $\small{\textcircled{1}}$ follows from smoothness of the objective function. 
 Dividing both sides by $\frac{\mu}{2}$, gives the desired bound of 
 $$ \|\hat{\theta}_{n,-U}(\lambda) - \tilde{\theta}_{n,-U}(\lambda)\|_2 
 \leq \frac{2m^2 C L}{\mu^{2} n^2} + \frac{Mm^2 L^2}{ \mu^3 n^2}$$
 For the non-smooth version of our algorithm, the same proof holds where we define 
 \begin{itemize}
  \item $\psi_{1} = \tilde{\loss}_{n,-U}(z, \theta, \lambda)$
     \item $\psi_{2} = \langle \nabla \tilde{\loss}_{n, -U}(\estt), \estt - \theta \rangle + \langle \estt - \theta, \nabla_{\theta}^{2}\tilde{\loss}_{n, -U}(\estt)[\estt - \theta]\rangle + \pi(\theta)$
    \item $\psi_{3} = \langle \nabla \tilde{\loss}_{n, -U}(\estt), \estt - \theta \rangle + \langle \estt - \theta, \nabla_{\theta}^{2}\tilde{\loss}_n(\estt)[\estt - \theta]\rangle + \pi(\theta)$
        \item $\hat{\theta}_{n,-U}(\lambda) = \argmin \psi_{1}(\theta)$, 
    \item $\Tilde{\theta}_{n,-U}(\lambda) = \argmin\psi_{3}(\theta)$
 \end{itemize}
 \begin{talign*}
 \frac{\mu}{2}\|\hat{\theta}_{n,-U}(\lambda) - \tilde{\theta}_{n,-U}(\lambda) \|_2 & \leq \|\nabla_{\theta}^{2}\tilde{\loss}_{n}(\hat{\theta}_{n,-U}(\lambda)) - \nabla_{\theta}^{2}\tilde{\loss}_{n,-U}(\hat{\theta}_{n,-U}(\lambda))\|_2\|\estt - \hat{\theta}_{n,-U}(\lambda)\|_2\\
&\quad + \|\nabla_\theta\psi_{2}(\hat{\theta}_{n,-U}(\lambda))  - \nabla_\theta\psi_{1}(\hat{\theta}_{n,-U}(\lambda)) \|_2 \\
& \leq \frac{m^2 C L}{\mu n^2} + \|\nabla_\theta \psi_{2}(\hat{\theta}_{n,-U}(\lambda))  - \nabla_\theta\psi_{1}(\hat{\theta}_{n,-U}(\lambda)) \|_2 \\
& \leq \frac{m^2 C L}{\mu n^2} + \frac{M}{2}\|\hat{\theta}_{n,-U}(\lambda) - \estt\|_2^2 \\
& \leq \frac{m^2 C L}{\mu n^2} + \frac{M}{2}\cdot \frac{m^2 L^2}{\mu^2 n^2} 
 \end{talign*}
 \subsection{Comparisons between batch and streaming algorithm}\label{app:batch_eq_stream}
 We show that the batch and streaming version of the algorithms are equivalent.
\paragraph{Case 1: $\pi$ is smooth.} The bounds we have proved are for the minizmier of $\varphi_3$, namely 
 \begin{talign*}
 \tilde{\theta}_{n,-U}(\lambda) &= \estt - \nabla_{\theta}^2 \tilde{F}(\estt)^{-1} \nabla \tilde{F}_{n,-U}(\estt)\\
  &=  \estt + \frac{1}{n}(\frac{1}{n}\sum_{i=1}^n\nabla_{\theta}^2 F(z_i,\estt, \lambda))^{-1} \sum_{i\in U}\nabla \ell(z_i, \estt)
 \end{talign*}
Now suppose 1 datapoint (user $j$) requests to be deleted. Then the streaming and batch algorithms agree, as the update becomes 
 \begin{talign*}
 \tilde{\theta}_{n,-i}(\lambda) &= \estt + \frac{1}{n}(\frac{1}{n}\sum_{i=1}^n\nabla_{\theta}^2 F(z_i,\estt, \lambda))^{-1} \nabla \ell(z_i, \estt).
   \end{talign*}
   Now suppose the algorithms are consistent for all deletion requests in the set $U$. When an additional user $j$ requests to delete their data the streaming algorithm returns \begin{talign*}
    \tilde{\theta}_{n,-(U \cup \{j\})}(\lambda) &=  \tilde{\theta}_{n,-U}(\lambda) + \frac{1}{n}(\frac{1}{n}\sum_{i=1}^n\nabla_{\theta}^2 F(z_i,\estt, \lambda))^{-1} \nabla \ell(z_j, \estt)\\
    & =  \estt + \frac{1}{n}(\frac{1}{n}\sum_{i=1}^n\nabla_{\theta}^2 F(z_i,\estt, \lambda))^{-1} \sum_{i\in U} \nabla \ell(z_i, \estt)\\&+  \frac{1}{n}(\frac{1}{n}\sum_{i=1}^n\nabla_{\theta}^2 F(z_i,\estt, \lambda))^{-1} \nabla \ell(z_j, \estt)\\
    & =  \estt +  \frac{1}{n}(\frac{1}{n}\sum_{i=1}^n\nabla_{\theta}^2 F(z_i,\estt, \lambda))^{-1} \nabla \sum_{i \in (U \cup \{j\})}\ell(z_i, \estt)
   \end{talign*}
   which matches the batch version of the deletion algorithm. This inductive arguments show both batch and streaming algorithms are the same. 
\paragraph{Case 2: $\pi$ is not smooth.} When $\pi$ is not smooth, the minimizer of $\varphi_3$ satisfies
  \begin{talign*}
  \tilde{\theta}_{n,-(U \cup \{j\})}(\lambda) &=  \tilde{\theta}_{n,-U}(\lambda) + \frac{1}{n}(\frac{1}{n}\sum_{i=1}^n\nabla_{\theta}^2 F(z_i,\estt, \lambda))^{-1} \nabla \ell(z_j, \estt) +\lambda \nabla \pi(    \tilde{\theta}_{n,-(U \cup \{j\})}(\lambda) )
  \end{talign*}
  When 1 datapoint (user $j$) requests to be deleted, the streaming and batch algorithms agree given $U = \emptyset.$ Now suppose the algorithms are consistent for all deletion requests in the set $U$. When an additional user $j$ requests to delete their data the streaming algorithm returns an estimator that satisfies
  \begin{talign*}
      \bar{\theta}_{n,-(U \cup \{j\})}(\lambda) &=  \bar{\theta}_{n,-U}(\lambda) + \frac{1}{n}H_{\ell}^{-1} \nabla \ell(z_j, \estt) + \lambda H_{\ell}^{-1}\nabla (\bar{\theta}_{n,-(U \cup \{j\})}(\lambda)) \\
       & =  \estt +  \frac{1}{n}H_\ell^{-1} \nabla \sum_{i \in (U \cup \{j\})}\ell(z_i, \estt)+ \lambda H_{\ell}^{-1}\nabla (\bar{\theta}_{n,-(U \cup \{j\})}(\lambda))
  \end{talign*}
    which matches the batch version of the deletion algorithm. This inductive arguments show both batch and streaming algorithms are the same. 

\subsection{Proof of excess empirical risk}
Second, we prove the excess empirical risk of our unlearning algorithm~(\ref{algo:1}).

\begin{proof}
\begin{talign*}
\mathbb{E}[F_{n} (\Tilde{\theta}_{n, -U}(\lambda)) - F_{n}(\theta^{*}(\lambda))] &= \mathbb{E}[F_{n} (\Tilde{\theta}_{n, -U}(\lambda)) - F_{n} (\hat{\theta}_{n}(\lambda)) + F_{n} (\hat{\theta}_{n}(\lambda)) - F_{n}(\theta^{*}(\lambda))] \\ 
& = \mathbb{E}[F_{n} (\Tilde{\theta}_{n, -U}(\lambda)) - F_{n} (\hat{\theta}_{n}(\lambda))] + \mathbb{E}[F_{n} (\hat{\theta}_{n}(\lambda)) - F_{n}(\theta^{*}(\lambda))] \\
& \overset{\tiny{\textcircled{1}}}{\leq} \mathbb{E}[L\|\Tilde{\theta}_{n, -U}(\lambda) - \hat{\theta}_{n}(\lambda)\|] + \frac{4L^{2}}{\mu n} \\
\end{talign*}

where $\textcircled{1}$ comes from Lemma~\ref{lemma:shalev} given that $F_n$ satisfies Assumption~\ref{ass:1} or~\ref{ass:2}.

Next we upper bound $\mathbb{E}[\|\Tilde{\theta}_{n, -U}(\lambda) - \hat{\theta}_{n}(\lambda)\|]$:

\begin{talign*}
\mathbb{E}[\|\Tilde{\theta}_{n, -U}(\lambda) - \hat{\theta}_{n}(\lambda)\|] &= \mathbb{E}[\|\Tilde{\theta}_{n, -U}(\lambda) - \hat{\theta}_{n, -U}(\lambda) + \hat{\theta}_{n, -U}(\lambda) - \hat{\theta}_{n}(\lambda)\|] \\
&= \mathbb{E}[\|\Tilde{\theta}_{n, -U}(\lambda) - \hat{\theta}_{n, -U}(\lambda)\|] + \mathbb{E}[\|\hat{\theta}_{n, -U}(\lambda) - \hat{\theta}_{n}(\lambda)\|] \\
& \overset{\tiny{\textcircled{2}}}{\leq} \mathbb{E}[\|\Tilde{\theta}_{n, -U}(\lambda) - \hat{\theta}_{n, -U}(\lambda)\|] + \frac{mL}{\mu n} \\
& \leq \mathbb{E}[\|\bar{\theta}_{n, -U}(\lambda) - \hat{\theta}_{n, -U}(\lambda) + \sigma\|] + \frac{mL}{\mu n} \\
& \leq \mathbb{E}[\|\bar{\theta}_{n, -U}(\lambda) - \hat{\theta}_{n, -U}(\lambda)\|] + \mathbb{E}[\|\sigma\|] + \frac{mL}{\mu n}\\
& \overset{\tiny{\textcircled{3}}}{\leq} \frac{2m^2 C L}{\mu^{2} n^2} + \frac{Mm^2 L^2}{ \mu^3 n^2} + \sqrt{d}c + \frac{mL}{\mu n} \\
& \leq \frac{2m^2 C L}{\mu^{2} n^2} + \frac{Mm^2 L^2}{ \mu^3 n^2} + \frac{\sqrt{d}\sqrt{2 \text{ln}(1.25/\delta)}}{\epsilon}( \frac{2m^2 C L}{\mu^{2} n^2} + \frac{Mm^2 L^2}{ \mu^3 n^2}) + \frac{mL}{\mu n}
\end{talign*}

where $\textcircled{2}$ comes from Lemma~\ref{lemma:noise} and $\textcircled{3}$ comes from Jensen's inequality and Lemma~\ref{lemma:noise} (Equation \ref{eq:ineq1}).

Now we substitute this back into our earlier bound:

\begin{talign*}
\mathbb{E}[F_{n} (\Tilde{\theta}_{n, -U}(\lambda)) - F_{n}(\theta^{*}(\lambda))] 
& \leq L(\frac{2m^2 C L}{\mu^{2} n^2} + \frac{Mm^2 L^2}{ \mu^3 n^2} + \frac{\sqrt{d}\sqrt{2 \text{ln}(1.25/\delta)}}{\epsilon}( \frac{2m^2 C L}{\mu^{2} n^2} + \frac{Mm^2 L^2}{ \mu^3 n^2}) + \frac{mL}{\mu n}) + \frac{4L^{2}}{\mu n} \\
& \leq \frac{2m^2 C L^{2}}{\mu^{2} n^2} + \frac{Mm^2 L^3}{ \mu^3 n^2} + \frac{\sqrt{d}\sqrt{2 \text{ln}(1.25/\delta)}}{\epsilon}( \frac{2m^2 C L^{2}}{\mu^{2} n^2} + \frac{Mm^2 L^3 }{ \mu^3 n^2}) + \frac{mL^{2}}{\mu n}) + \frac{4L^{2}}{\mu n} \\
& \leq (1 + \frac{\sqrt{d}\sqrt{2 \text{ln}(1.25/\delta)}}{\epsilon})(\frac{2m^2 C L^{2}}{\mu^{2} n^2} + \frac{Mm^2 L^3}{ \mu^3 n^2}) + \frac{4mL^2}{\mu n} \\
& \leq (1 + \frac{\sqrt{d}\sqrt{2 \text{ln}(1.25/\delta)}}{\epsilon})(\frac{(2C\mu + ML)m^2 L^2}{\mu^3 n^2}) + \frac{4mL^2}{\mu n}
\end{talign*}

\end{proof}

Finally, we prove that our unlearning algorithm (\ref{algo:1}) results in $(\epsilon, \delta)$-certifiable removal of datapoint $\mathbf{z} \in U \subseteq S$.

\begin{proof}
We use a similar technique to the proof of the differential privacy guarantee for the Gaussian mechanism (\cite{dwork2014algorithmic}).

Let $\hat{\theta}_{n}(\lambda)$ be the output of learning algorithm $A$ trained on dataset $S$ and $\tilde{\theta}_{n, -U}(\lambda)$ be the output of unlearning algorithm $M$ run on the sequence of delete requests $U$, $\hat{\theta}_{n}(\lambda)$, and the data statistics $T(S)$. We also note the output of $M$ before adding noise as $\bar{\theta}_{n, -U}(\lambda)$. Finally, we denote $\hat{\theta}_{n,-U}(\lambda)$ as the output of $A$ trained on the dataset $S \backslash U$.

We note that in~\cref{algo:1} that $\tilde{\theta}_{n, -U}(\lambda)$ is simply $\tilde{\theta}_{n, -U}(\lambda) = \bar{\theta}_{n, -U}(\lambda) + \sigma$. The noise $\sigma$ is sampled from $\mathcal{N}(0, c^{2}I)$ with $c = \|\hat{\theta}_{n, -U}(\lambda) - \bar{\theta}_{n,-U}(\lambda)\|_{2} \cdot \frac{\sqrt{2\text{ln}(1.25/\delta)}}{\epsilon}$. Where $ \|\hat{\theta}_{n, -U}(\lambda) - \bar{\theta}_{n,-U}(\lambda)\|_{2} \leq \frac{2m^2 CL}{n^2\mu^2 } + \frac{m^2 ML^2}{n^2 \mu^3}$ (\ref{eq:ineq1}). Following the same proof for the DP gaurantee of the Gaussian mechanism as~\citet{dwork2014algorithmic} (Theorem A.1) given the noise is sampled from the previously described Gaussian distribution we get for any $\Theta$:

\begin{align*}
 P(\hat{\theta}_{n,-U} \in \Theta)&\leq e^{\epsilon} P(\tilde{\theta}_{n,-U} \in \Theta) + \delta, \quad \text{and}\\
P(\tilde{\theta}_{n,-U} \in \Theta) &\leq e^{\epsilon} P(\hat{\theta}_{n,-U} \in \Theta)+ \delta
\end{align*}

resulting in $(\epsilon, \delta)$-unlearning.

% \begin{talign*}

% \end{talign*}
\end{proof}

\section{Proof of Algorithm~\ref{algo:1} 
Deletion Capacity}\label{app:del_cap}

The upper bound on the excess risk (Theorem~\ref{thm:1}) implies that we can delete at least:

    \begin{talign*}
    m_{\epsilon,\delta, \gamma}^{A,M}(d,n) \geq c \cdot \frac{n \sqrt{\epsilon}}{(d \text{log}(1/\delta))^{\frac{1}{4}}}
    \end{talign*}

where $c$ depends on the properties of function $F(z,\theta,\lambda)$. We specifically derive the value of $c$ below by substituting our deletion capacity bound as $m$ into the empirical excess risk upper bound:

\begin{talign}\label{eq:exrisk}
\mathbb{E}[F(\tilde{\theta}_{n, -U}(\lambda)) -  F(\theta^\ast(\lambda))] &= O\left( \frac{(2C\mu + ML) L^2m^2}{\mu^3n^2}\frac{\sqrt{d}\sqrt{ln(1/\delta)}}{\epsilon} + \frac{4mL^2}{\mu n}\right)  
\end{talign}
Plugging in the  deletion capacity bound $m = c \cdot \frac{n \sqrt{\epsilon}}{(d \text{log}(1/\delta))^{\frac{1}{4}}}$ into the excess risk bound~\eqref{eq:exrisk} then 
\begin{talign*}
 \frac{(2C\mu + ML) L^2m^2}{\mu^3n^2}\frac{\sqrt{d}\sqrt{ln(1/\delta)}}{\epsilon} + \frac{4mL^2}{\mu n} & =  
 %\gamma %
\frac{c^{2} (2C\mu + ML)L^{2}}{\mu^{3}} + \frac{4L^{2}c}{\mu}  \frac{n \sqrt{\epsilon}}{(d \text{log}(1/\delta))^{\frac{1}{4}}}\\
& \leq c \left(\frac{c (2C\mu + ML)L^{2}}{\mu^{3}} + \frac{4L^{2}}{\mu}\right)
\end{talign*}
Therefore,
\begin{talign*}
c \leq \gamma (\frac{\mu^{3}}{(2C\mu + ML)L^{2}} + \frac{\mu}{4L^{2}}) \qquad \Longrightarrow \qquad \mathbb{E}[F(\tilde{\theta}_{n, -U}(\lambda)) -  F(\theta^\ast(\lambda))] \leq \gamma
\end{talign*}
given $c \leq 1$. Note that the third line follows from the fact that $\frac{ \sqrt{\epsilon}}{(d \text{log}(1/\delta))^{\frac{1}{4}}} \leq 1$ given  $\epsilon \leq 1$ and $\delta \leq 0.005$.
\section{Extension of non-smooth regularizer to \cite{sekhari2021remember}}\label{app:ext} 

Given a function $F(z, \theta, \lambda)$ with a non-smooth regularizer $\pi(\theta)$ which satisfies 
Assumption~\ref{ass:2}, the algorithm from~\citet{sekhari2021remember} can use non-smooth regularizers with the same deletion capacity, generalization, and unlearning guarantees as~\cref{algo:1}. This follows from fact that the removal mechanism introduced by~\citet{sekhari2021remember} minimizes $\psi_2$ in~\cref{app:closeness_proof}. Therefore the optimizer comparison theorem can be applied and the distnace between the estimator and the leave-U-out estimator can be upper bounded by the same terms (more precisely, we can upper bound thist distance by $\frac{m^2ML^2}{n^2\mu^3}$). %the algorithm provided by~\citet{sekhari2021remember} being equivalent to~\cref{algo:1} in both the smooth and non-smooth settings which is given by~\cref{app:batch_eq_stream} Case 2.

\section{Dataset Details}\label{app:datasets}

\texttt{MNIST} We consider digit classification from the MNIST dataset which contains 60000 images of digits from 1-9. We select only digits 3 and 8 to simplify the task to binary classification. We flatten the original images which are $28 \times 28$ into a a vector of 784 pixels. Additionally, we allow for either random sampling or \textit{adaptive} sampling where the probability of sampling a 3 is set to 10\% and the probability of sampling an 8 is set to 90\%.

\texttt{SVHN} We consider digit recognition from street signs from the SVHN dataset which contains 60000 images of street sign images that contain digits from 1-9. We select only digits 3 and 8 to simplify the task to binary classification. We flatten the original images which are $28 \times 28$ into a a vector of 784 pixels. Additionally, we allow for either random sampling or \textit{adaptive} sampling where the probability of sampling a 3 is set to 10\% and the probability of sampling an 8 is set to 90\%.

\texttt{Warfarin Dosing} Warfarin is a prescription drug used to treat symptoms stemming from blood clots (e.g. deep vein thrombosis) and to help reduce the incidence of stroke and heart attack in at-risk patients. It is an anticoagulant which inhibits blood clotting but overdosing leads to excessive bleeding. The appropriate dosage for a patient dependent on demographic and physiologic factors resulting in high variance between patients. We focus on predicting small or large dosages for patients (defined as > 30mg/week) from a dataset released by the International Warfarin Pharmacogenetics Consortium~\cite{international2009estimation} which contains both demographic and physiological measurements for patients. The dataset contains 5528 examples each with 62 features.

\section{Additional Experiments}\label{app:exps}

\paragraph{Logistic Regression with Smooth Regularizers} We present the test accuracy results for the remaining values of $\lambda = \{10^{-4}, 10^{-5}, 10^{-6}\}$.

\begin{figure}[htb!]
\centering
\begin{subfigure}{.45\textwidth}
  \centering
  \includegraphics[width=\linewidth]{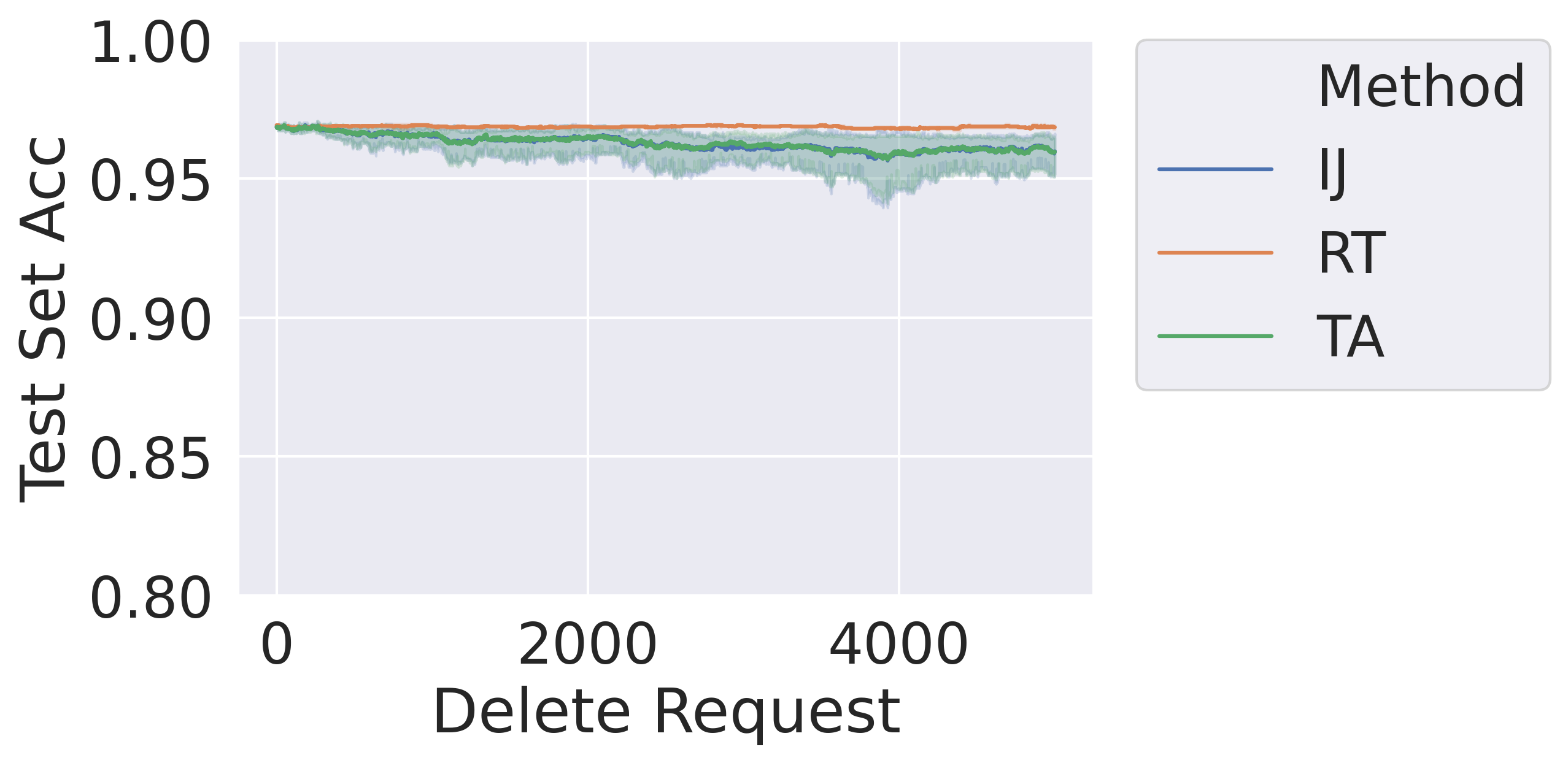}
  \label{fig:mnist_random_4}
\end{subfigure}%
\begin{subfigure}{.45\textwidth}
  \centering
  \includegraphics[width=\linewidth]{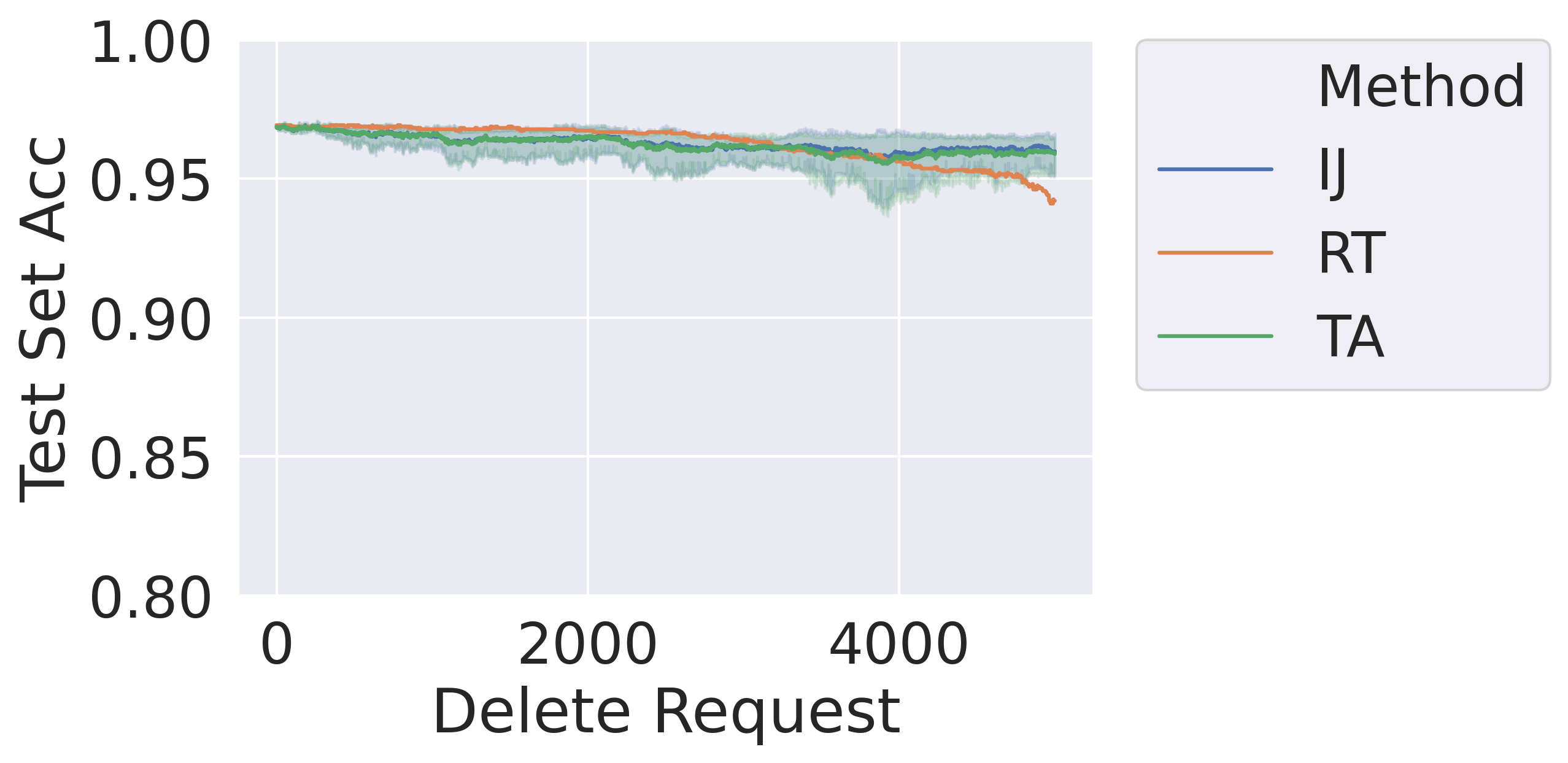}
  \label{fig:mnist_adversarial_4}
\end{subfigure}%
\caption{\textbf{IJ} vs. \textbf{RT} and \textbf{TA} for smooth regularizers. Comparing both the test accuracy of the unlearned models in our $\ell_2$ logistic regression setup for $\lambda = 10^{-4}$ for random vs adaptive sampling.}
\label{fig:mnist_4}
\end{figure}

\begin{figure}[htb!]
\centering
\begin{subfigure}{.45\textwidth}
  \centering
  \includegraphics[width=\linewidth]{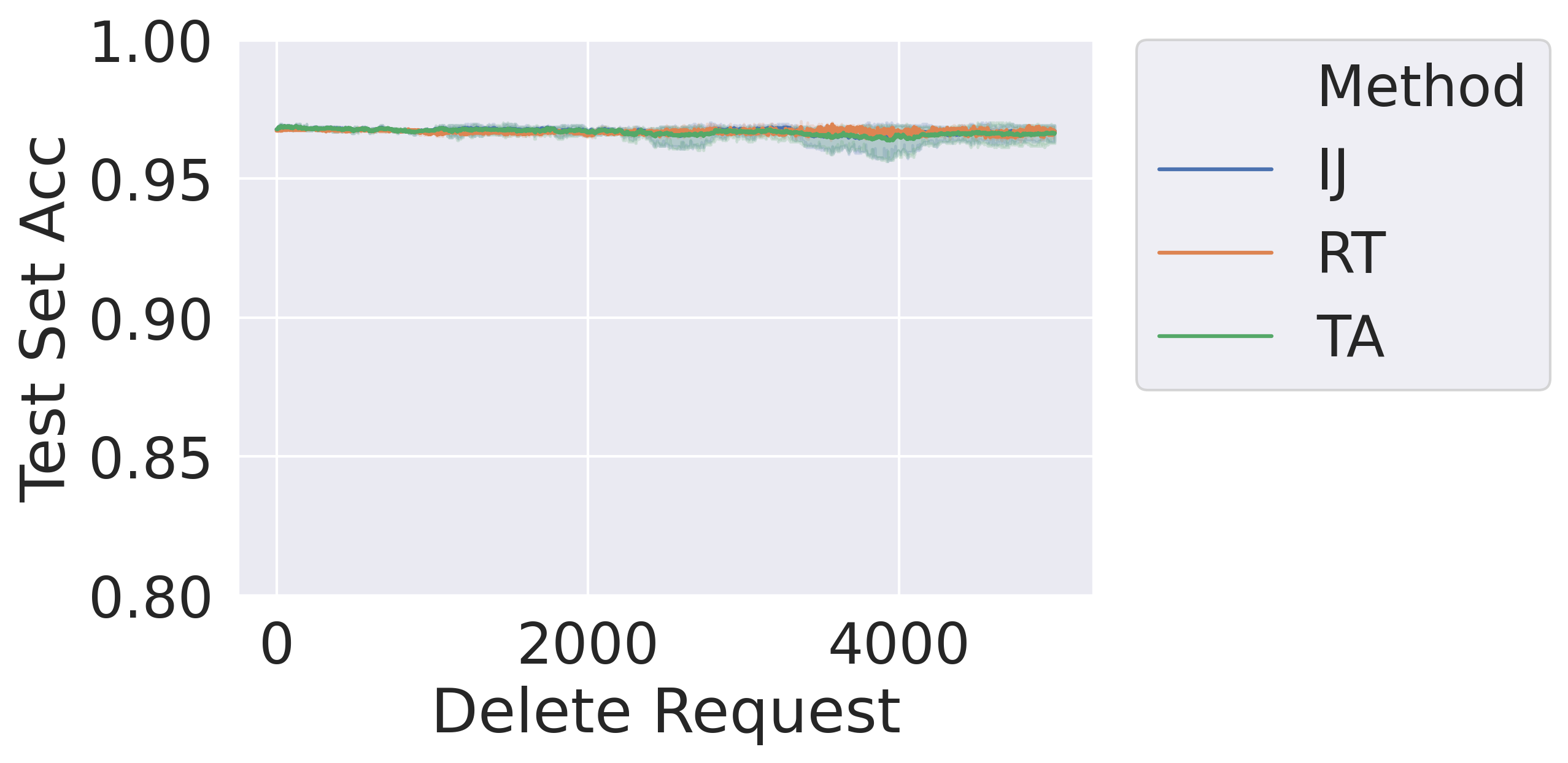}
  \label{fig:mnist_random_5}
\end{subfigure}%
\begin{subfigure}{.45\textwidth}
  \centering
  \includegraphics[width=\linewidth]{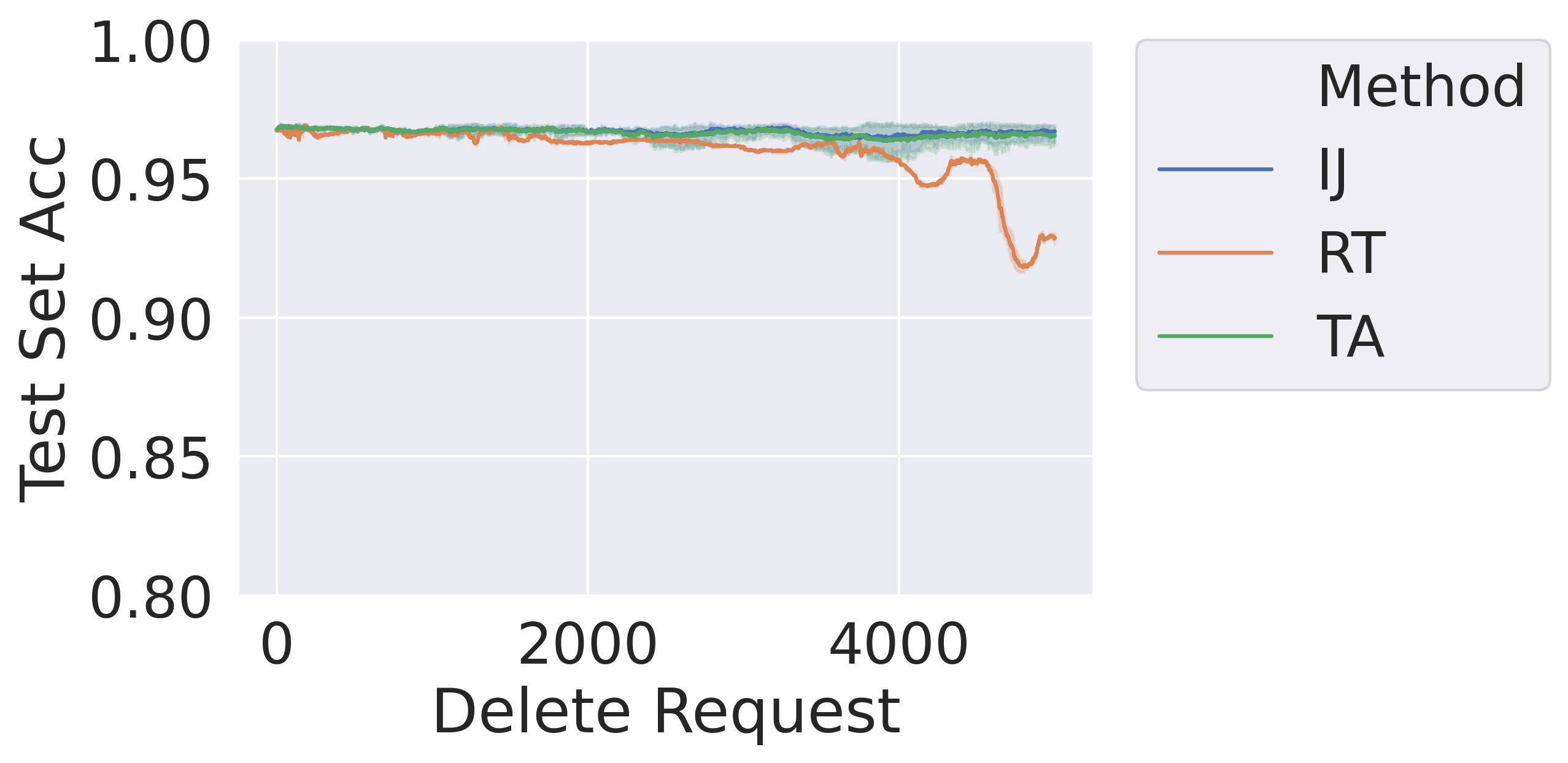}
  \label{fig:mnist_adversarial_5}
\end{subfigure}%
\caption{\textbf{IJ} vs. \textbf{RT} and \textbf{TA} for smooth regularizers. Comparing both the test accuracy of the unlearned models in our $\ell_2$ logistic regression setup for $\lambda = 10^{-5}$ for random vs adaptive sampling.}
\label{fig:mnist_5}
\end{figure}

\begin{figure}[htb!]
\centering
\begin{subfigure}{.45\textwidth}
  \centering
  \includegraphics[width=\linewidth]{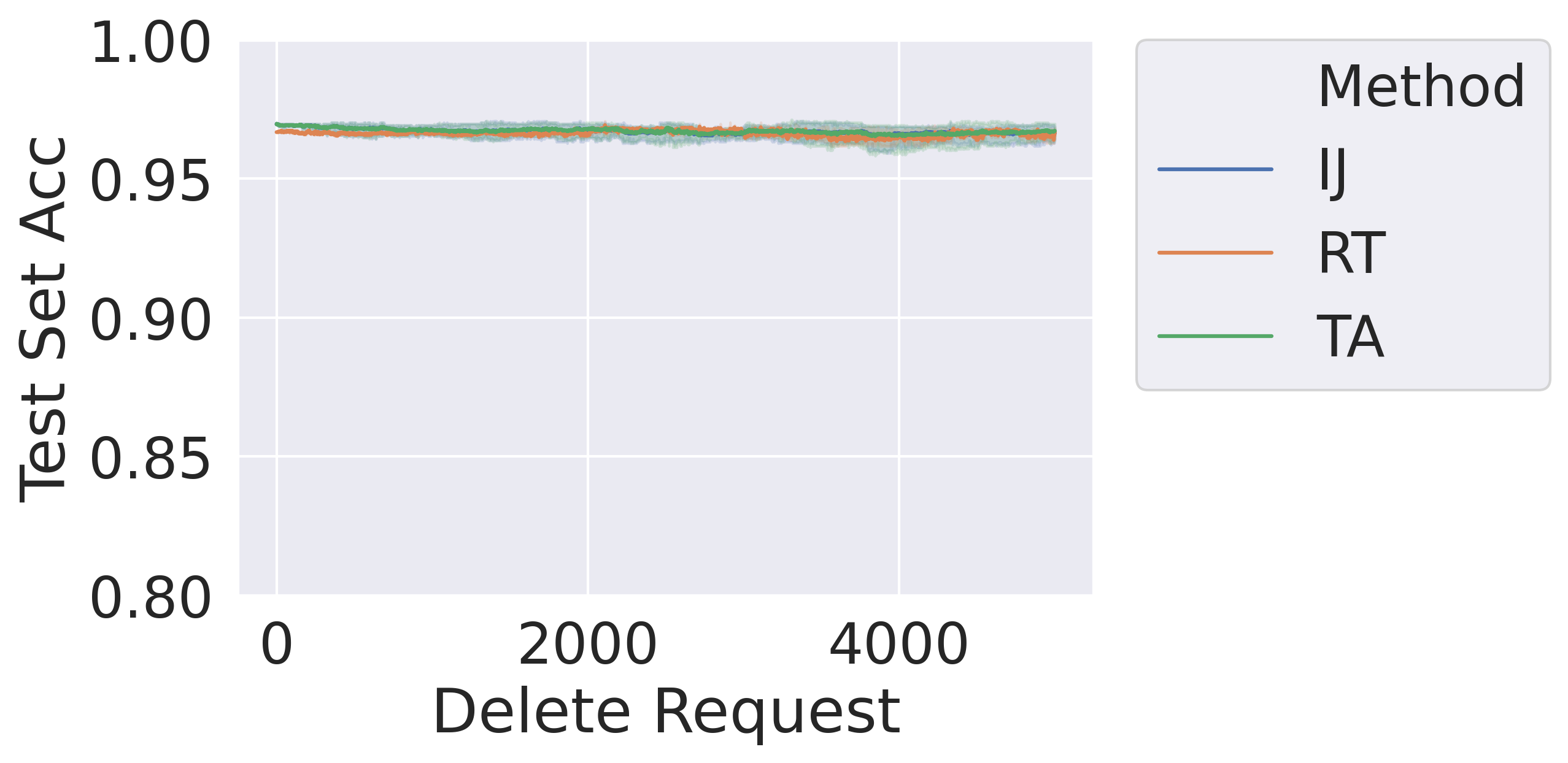}
  \label{fig:mnist_random_6}
\end{subfigure}%
\begin{subfigure}{.45\textwidth}
  \centering
  \includegraphics[width=\linewidth]{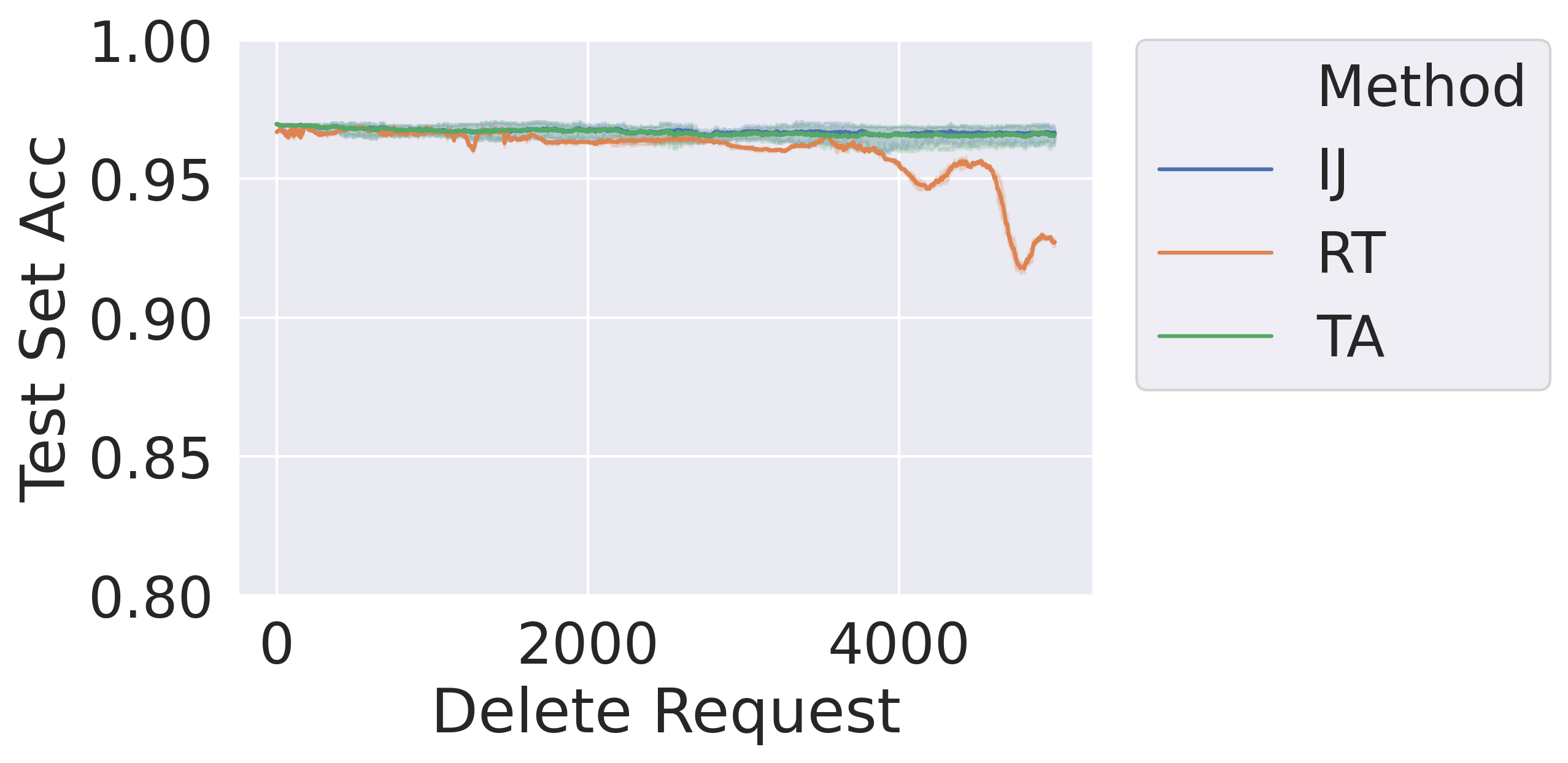}
  \label{fig:mnist_adversarial_6}
\end{subfigure}%
\caption{\textbf{IJ} vs. \textbf{RT} and \textbf{TA} for smooth regularizers. Comparing both the test accuracy of the unlearned models in our $\ell_2$ logistic regression setup for $\lambda = 10^{-6}$ for random vs adaptive sampling.}
\label{fig:mnist_6}
\end{figure}

\pagebreak
\paragraph{Logistic Regression with Non-Smooth Regularizers} We present the test accuracy results for the remaining values of $\lambda = \{10^{-4}, 10^{-5}, 10^{-6}\}$.

\begin{figure}[htb!]
\centering
\begin{subfigure}{.3\textwidth}
  \centering
  \includegraphics[width=\linewidth]{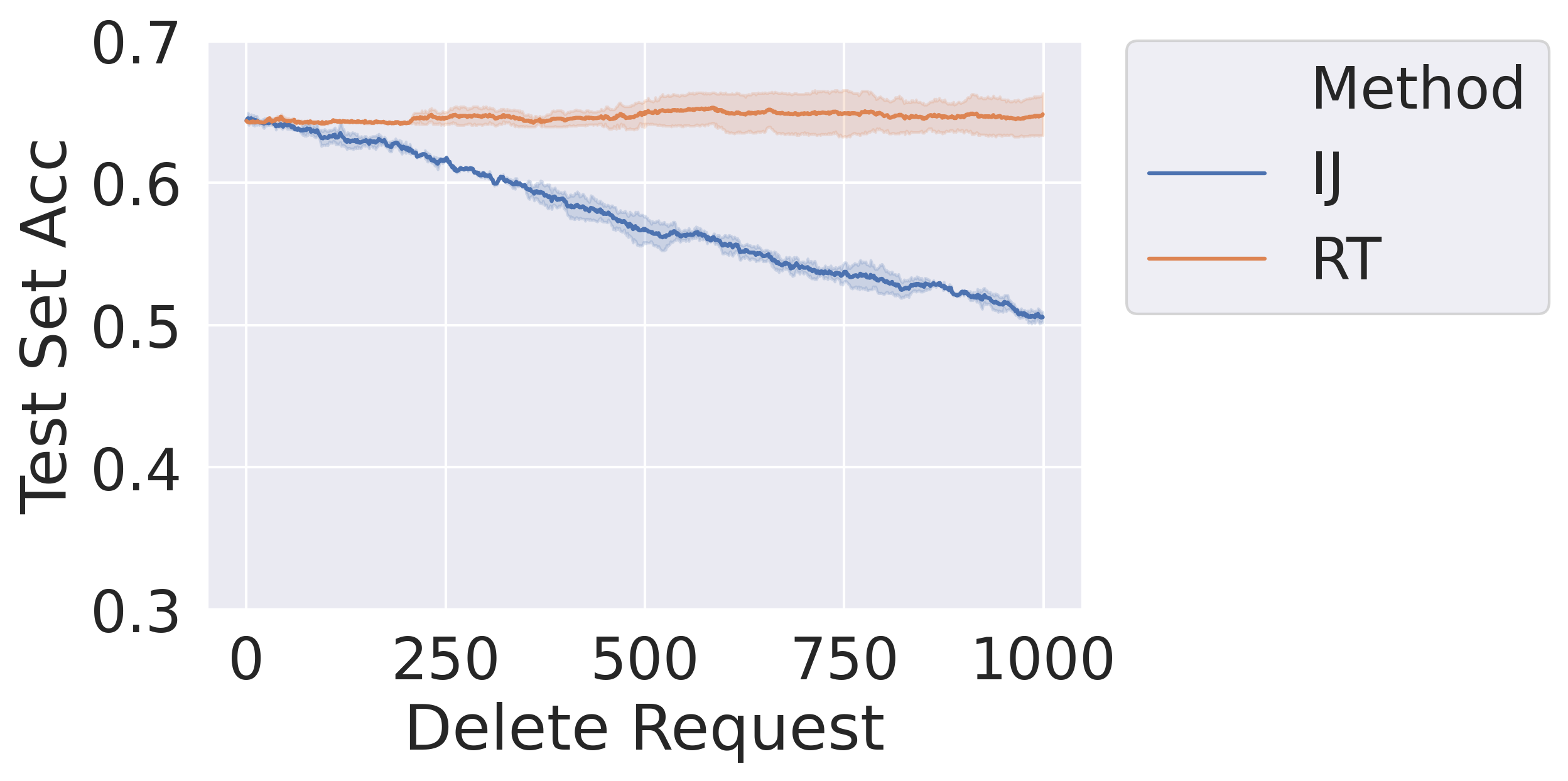}
  \label{fig:warfarin_random_4}
\end{subfigure}%
\begin{subfigure}{.3\textwidth}
  \centering
  \includegraphics[width=\linewidth]{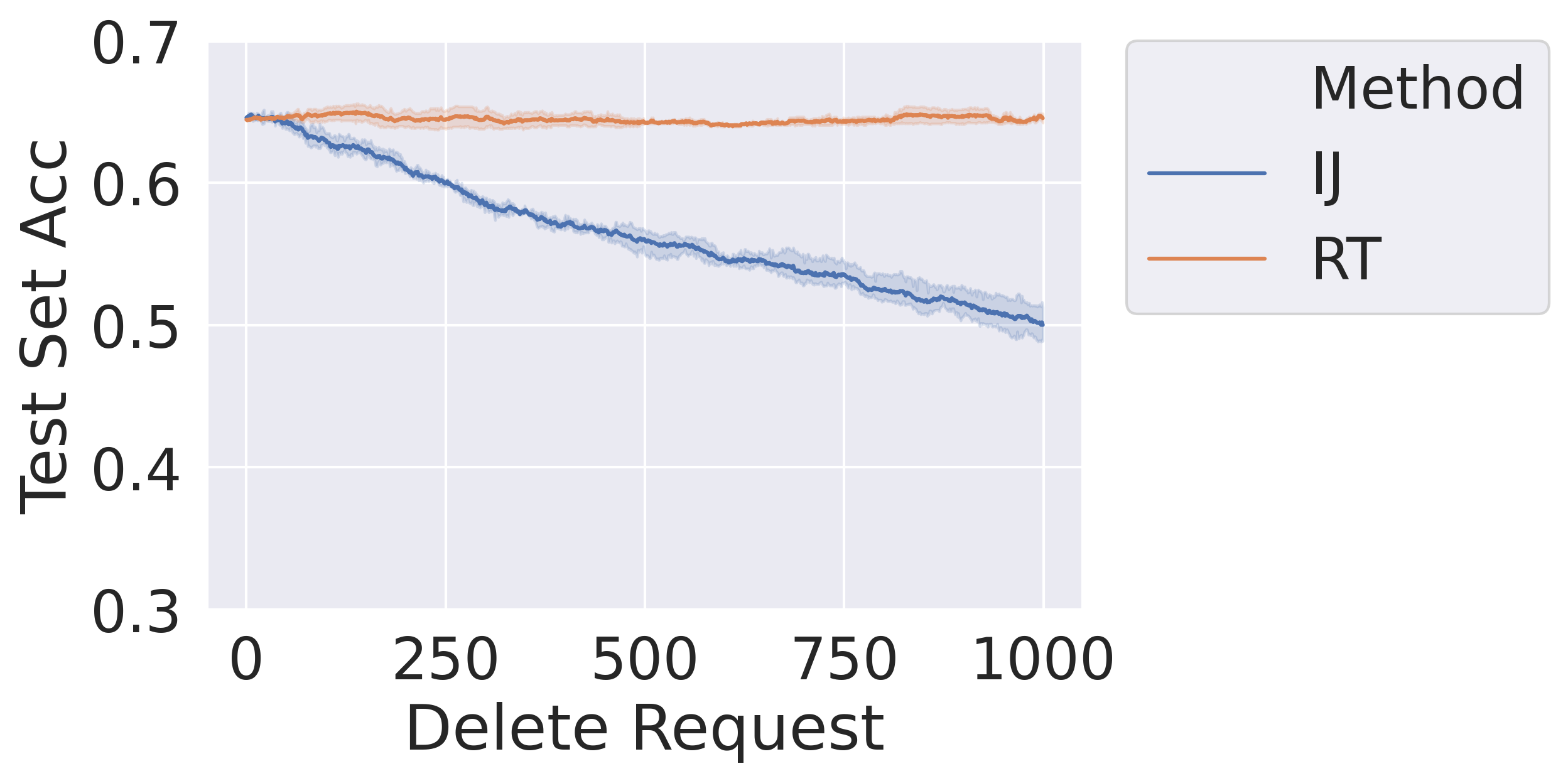}
  \label{fig:warfarin_random_5}
\end{subfigure}%
\begin{subfigure}{.3\textwidth}
  \centering
  \includegraphics[width=\linewidth]{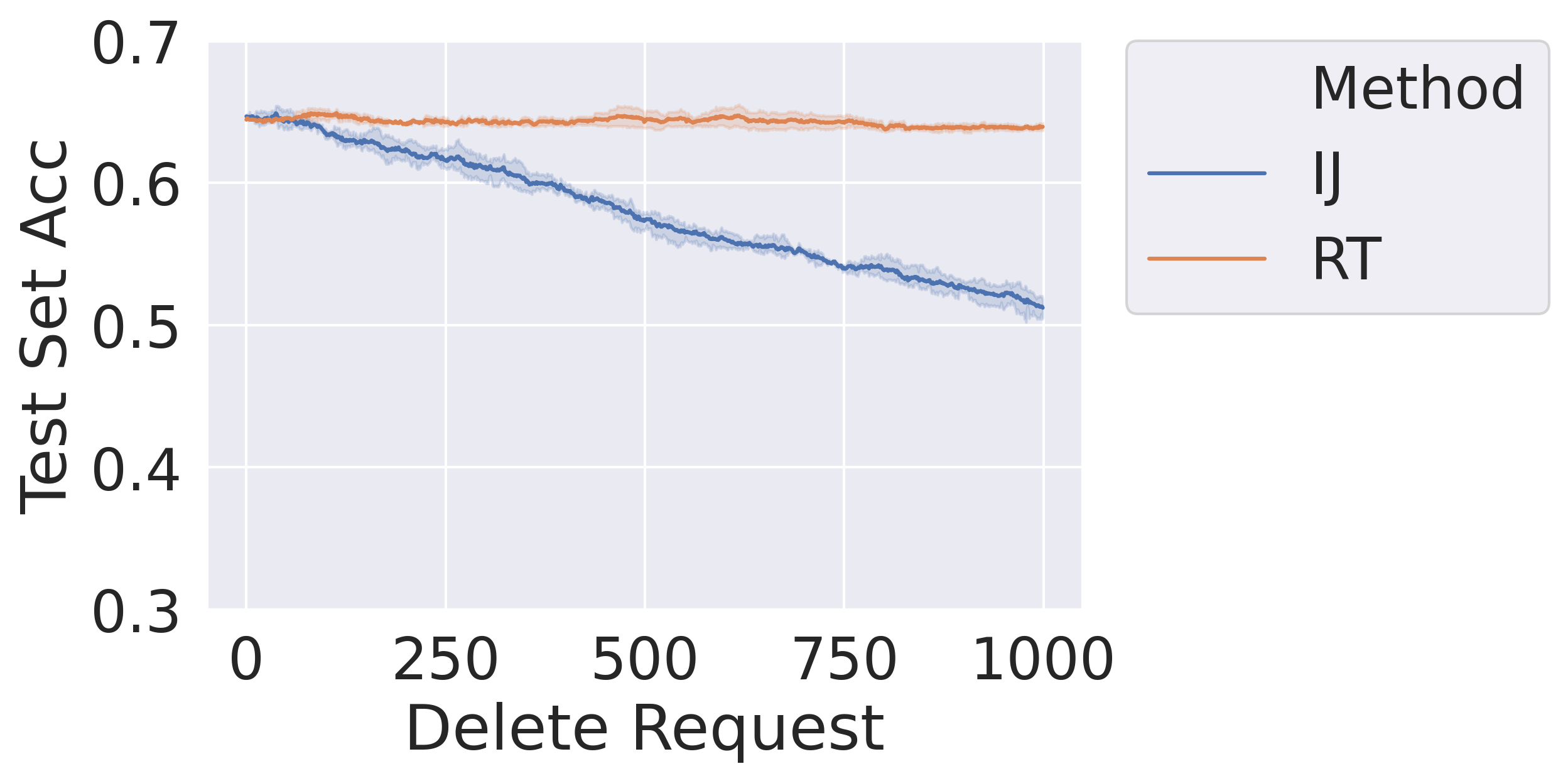}
  \label{fig:warfarin_random_6}
\end{subfigure}%
\caption{\textbf{IJ} vs. \textbf{RT} for non-smooth regularizers. Comparing the test accuracy of the unlearned models in our $\ell_1$ logistic regression setup for $\lambda \in \{10^{-4}, 10^{-5}, 10^{-6}\}$.}
\label{fig:wafarin_runs}
\end{figure}

\paragraph{Non-Conxex: Logistic Regression with Differentially Private Feature Extractor} We present the test accuracy results for the remaining values of $\lambda = \{10^{-4}, 10^{-5}, 10^{-6}\}$.

\begin{figure}[htb!]
\centering
\begin{subfigure}{.45\textwidth}
  \centering
  \includegraphics[width=\linewidth]{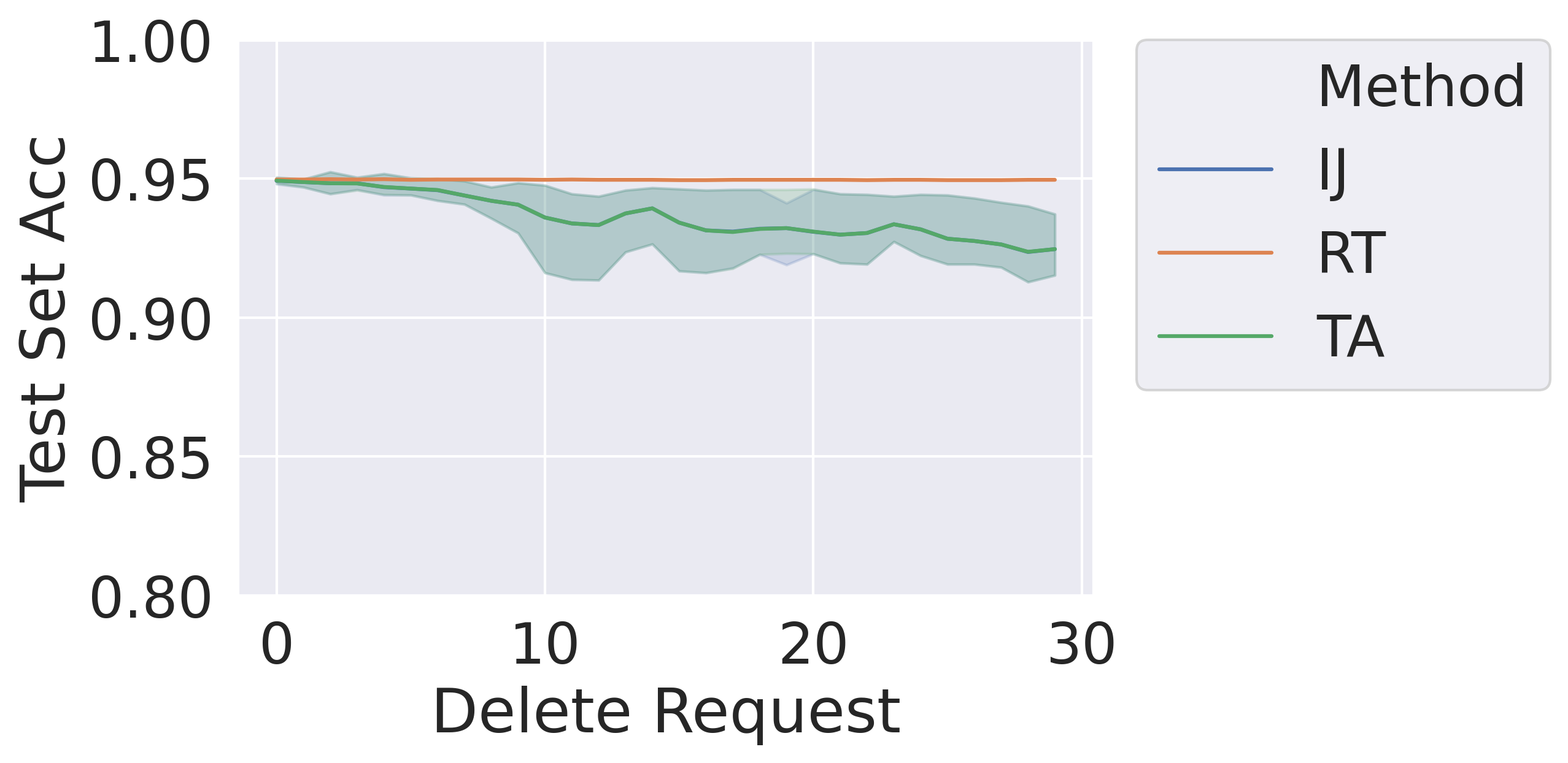}
  \label{fig:svhn_random_4}
\end{subfigure}%
\begin{subfigure}{.45\textwidth}
  \centering
  \includegraphics[width=\linewidth]{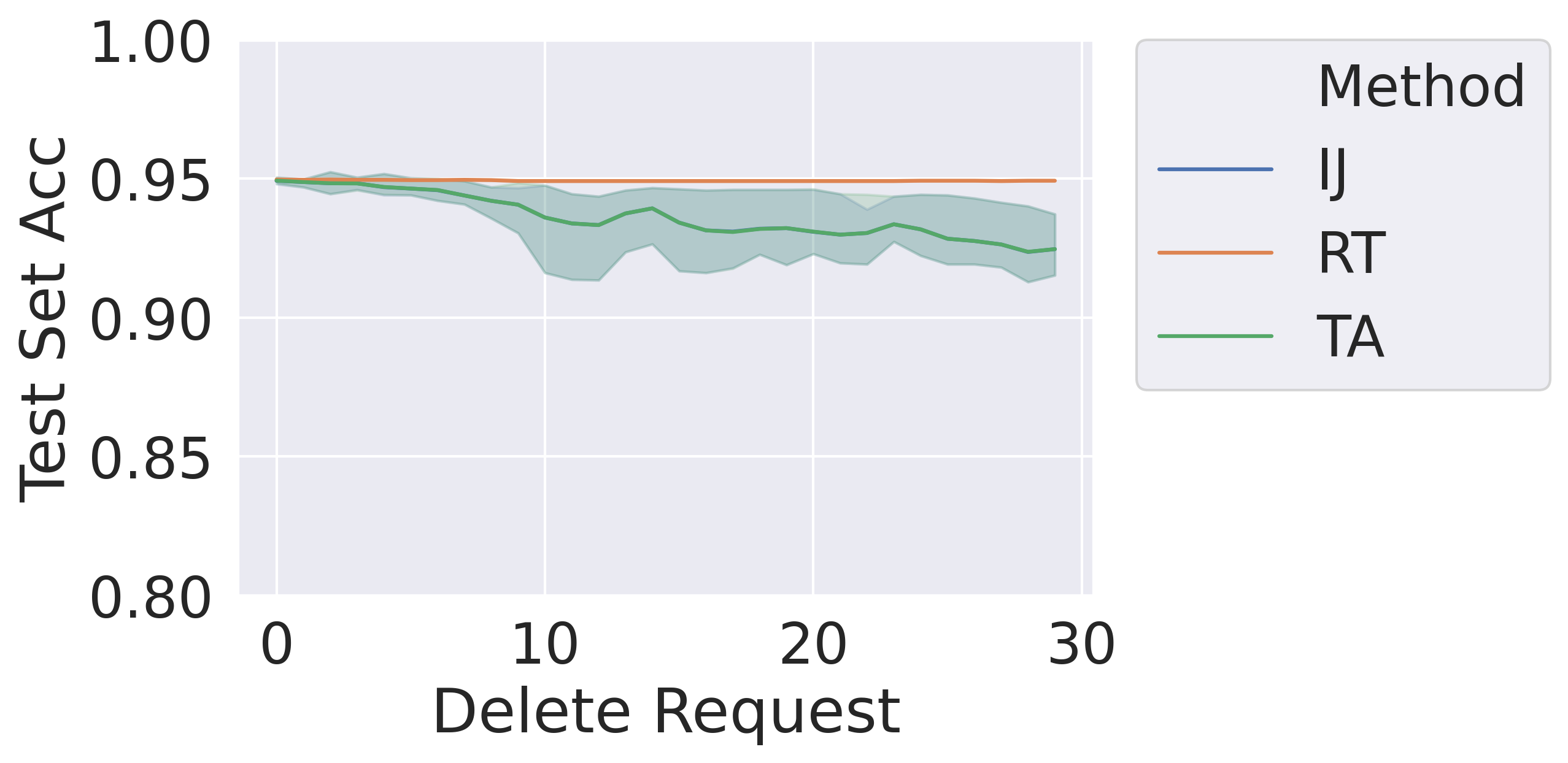}
  \label{fig:svhn_adversarial_4}
\end{subfigure}%
\caption{\textbf{IJ} vs. \textbf{TA} and \textbf{RT} for non-convex training. Comparing both the test accuracy of the unlearned models in our DP feature extractor + $\ell_2$ setup for $\lambda = 10^{-4}$.}
\label{fig:svhn_4}
\end{figure}

\begin{figure}[htb!]
\centering
\begin{subfigure}{.45\textwidth}
  \centering
  \includegraphics[width=\linewidth]{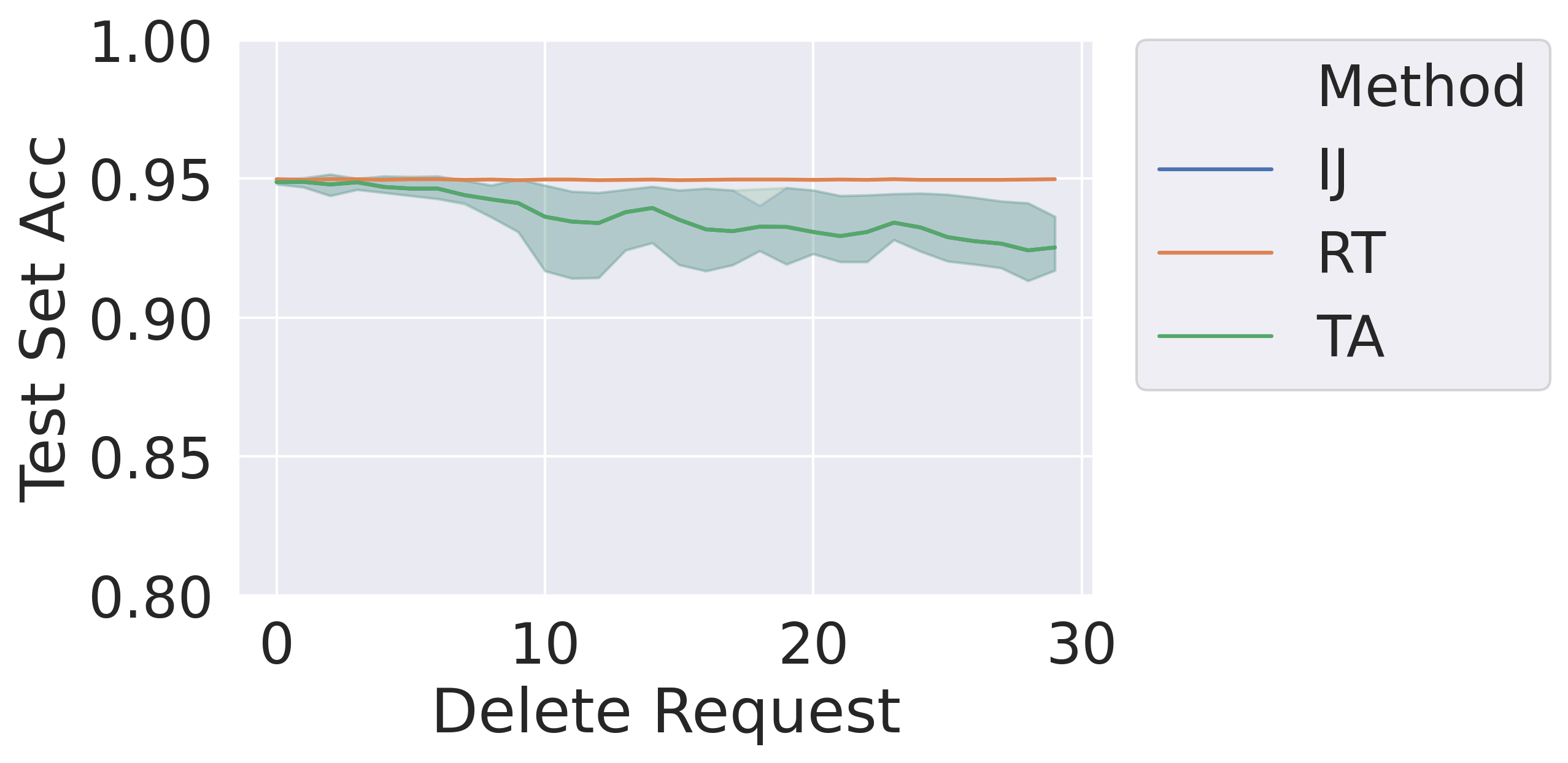}
  \label{fig:svhn_random_5}
\end{subfigure}%
\begin{subfigure}{.45\textwidth}
  \centering
  \includegraphics[width=\linewidth]{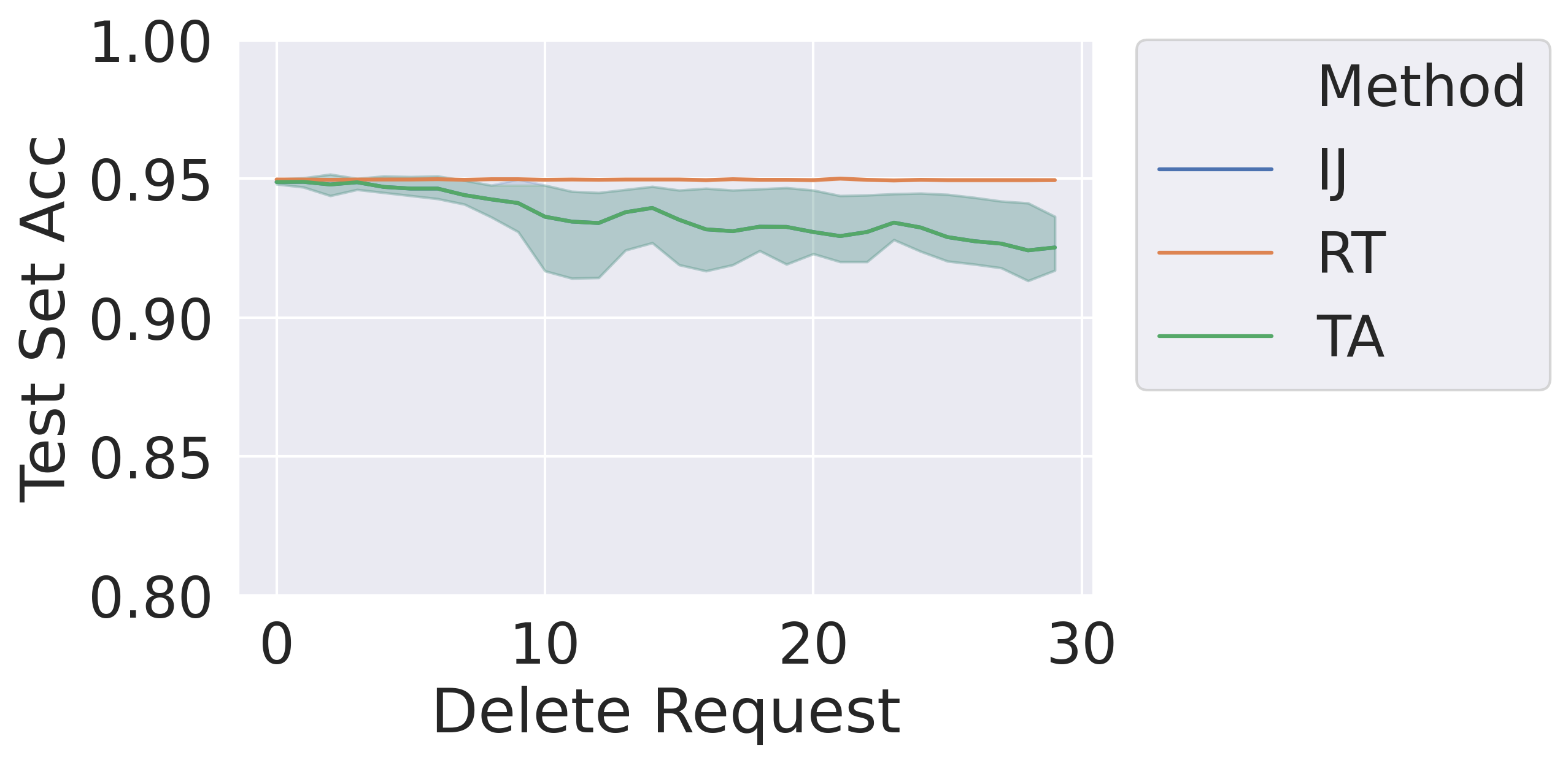}
  \label{fig:svhn_adversarial_5}
\end{subfigure}%
\caption{\textbf{IJ} vs. \textbf{TA} and \textbf{RT} for non-convex training. Comparing both the test accuracy of the unlearned models in our DP feature extractor + $\ell_2$ setup for $\lambda = 10^{-5}$.}
\label{fig:svhn_5}
\end{figure}

\begin{figure}[htb!]
\centering
\begin{subfigure}{.45\textwidth}
  \centering
  \includegraphics[width=\linewidth]{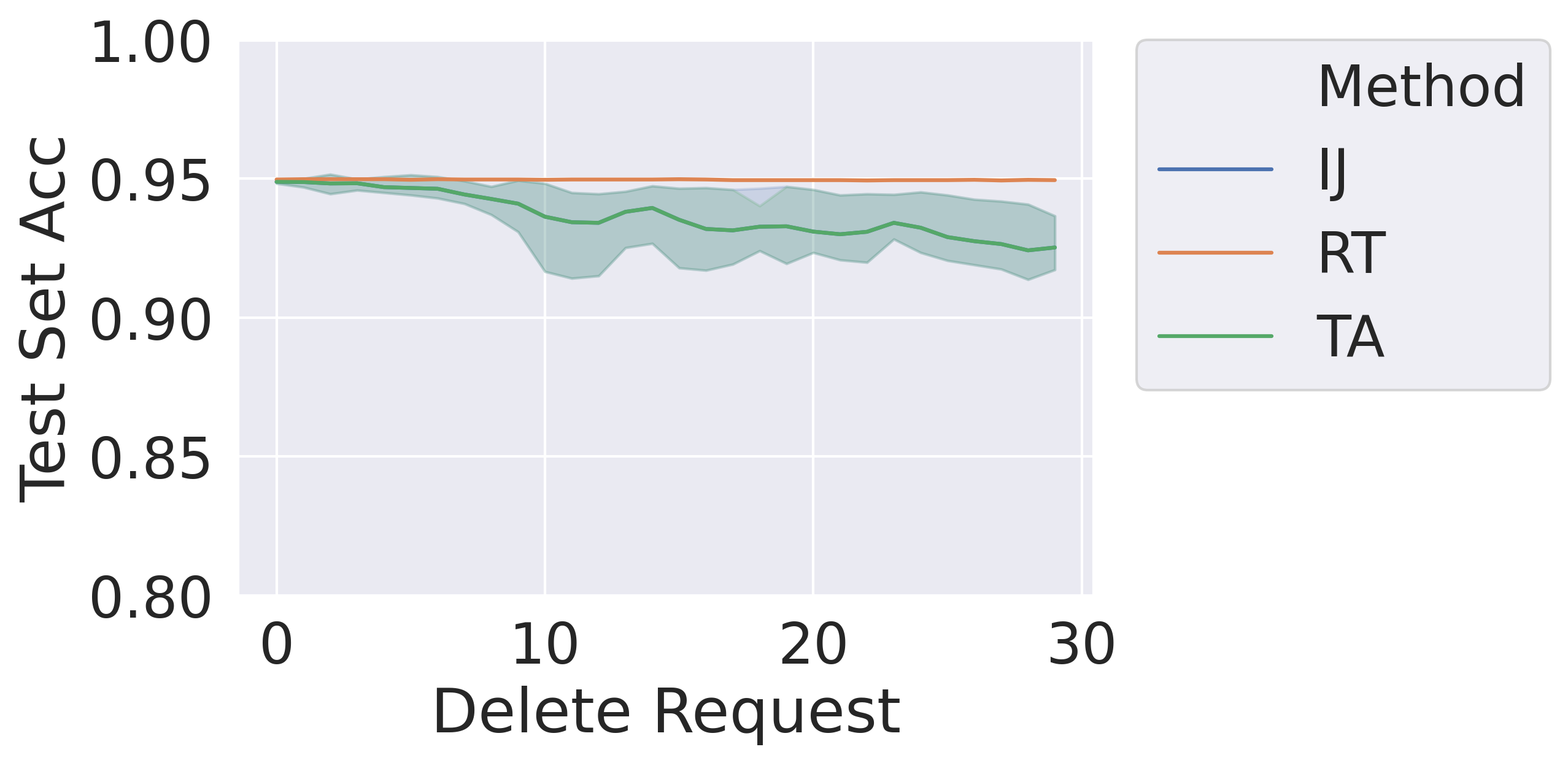}
  \label{fig:svhn_random_6}
\end{subfigure}%
\begin{subfigure}{.45\textwidth}
  \centering
  \includegraphics[width=\linewidth]{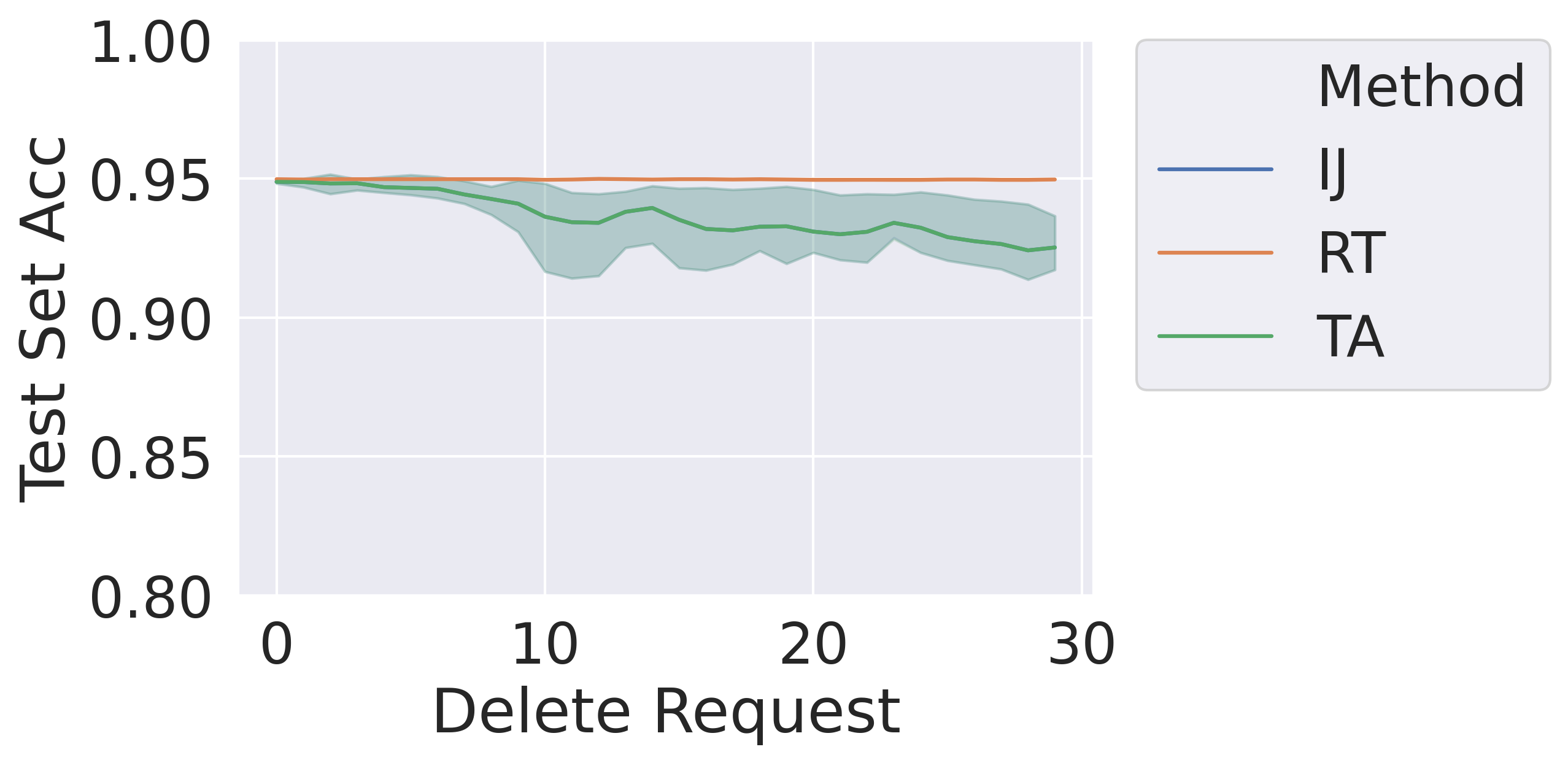}
  \label{fig:svhn_adversarial_6}
\end{subfigure}%
\caption{\textbf{IJ} vs. \textbf{TA} and \textbf{RT} for non-convex training. Comparing both the test accuracy of the unlearned models in our DP feature extractor + $\ell_2$ setup for $\lambda = 10^{-6}$.}
\label{fig:svhn_6}
\end{figure}

\pagebreak
\subsection{Runtimes}

\begin{figure}[htb!]
\centering
\begin{subfigure}{.3\textwidth}
  \centering
  \includegraphics[width=\linewidth]{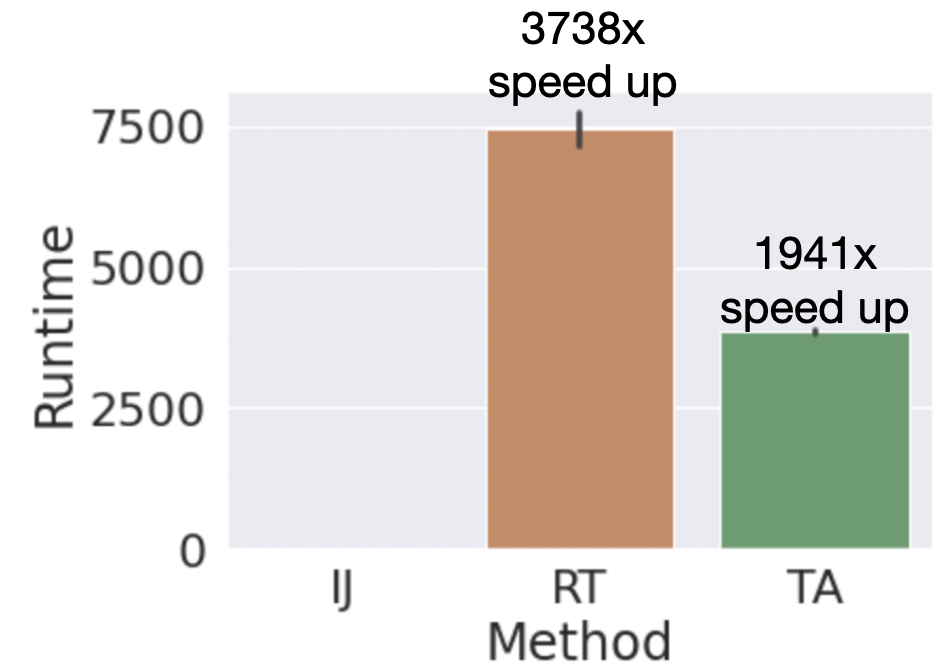}
  \label{fig:mnist_runtime_4}
\end{subfigure}%
\begin{subfigure}{.3\textwidth}
  \centering
  \includegraphics[width=\linewidth]{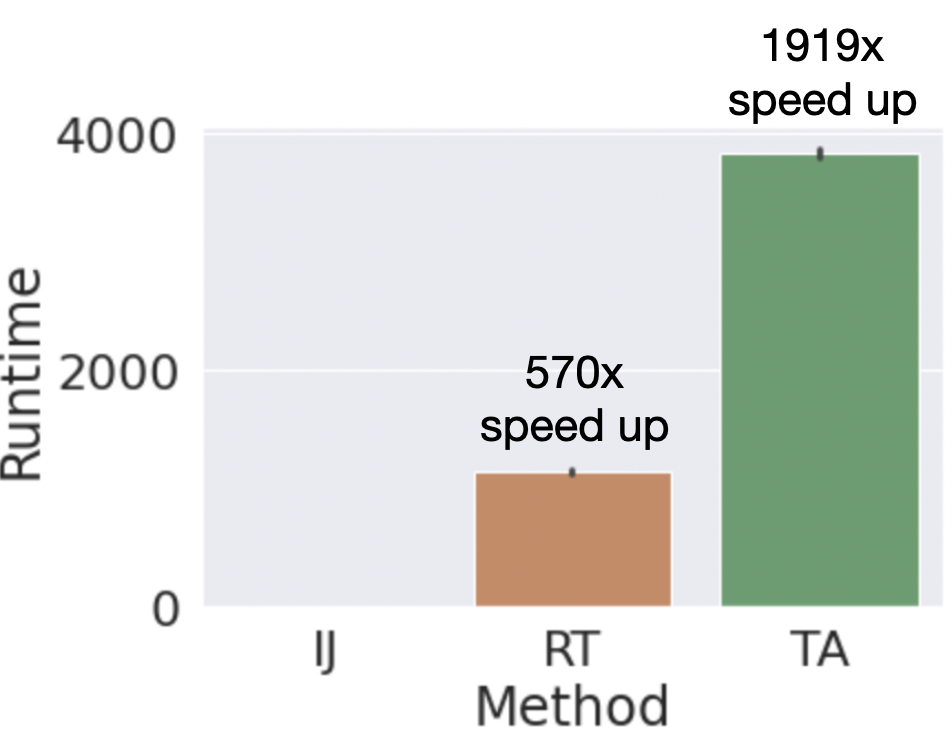}
  \label{fig:mnist_runtime_5}
\end{subfigure}%
\begin{subfigure}{.3\textwidth}
  \centering
  \includegraphics[width=\linewidth]{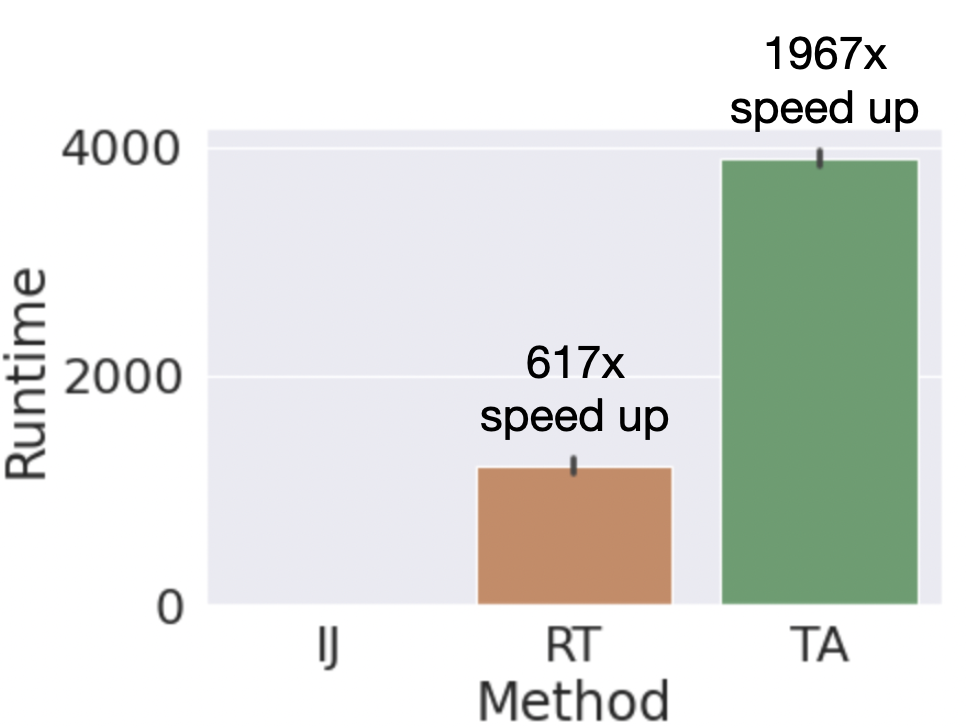}
  \label{fig:mnist_runtime_6}
\end{subfigure}%
\caption{\textbf{IJ} vs. \textbf{RT} vs. \textbf{TA} for smooth regularizers on MNIST. Demonstrating runtime improvements across different hyperparameter settings of $10^{-4}, 10^{-5}, 10^{-6}$.}
\label{fig:mnist_runtimes}
\end{figure}

\begin{figure}[htb!]
\centering
\begin{subfigure}{.3\textwidth}
  \centering
  \includegraphics[width=\linewidth]{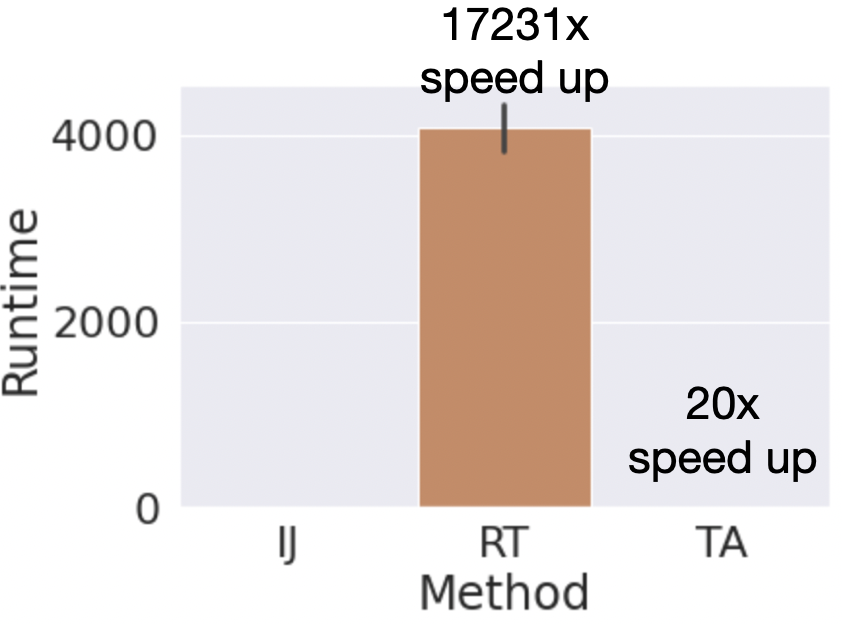}
  \label{fig:svhn_runtime_4}
\end{subfigure}%
\begin{subfigure}{.3\textwidth}
  \centering
  \includegraphics[width=\linewidth]{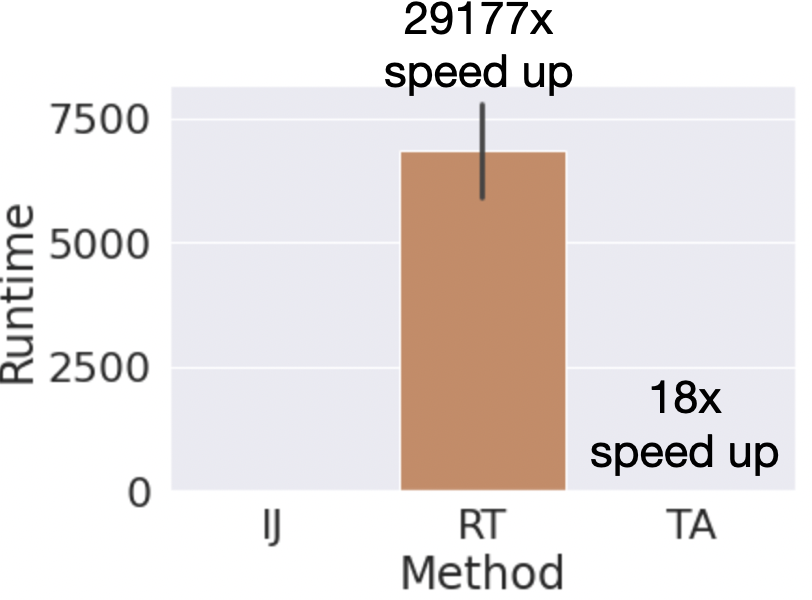}
  \label{fig:svhn_runtime_5}
\end{subfigure}%
\begin{subfigure}{.3\textwidth}
  \centering
  \includegraphics[width=\linewidth]{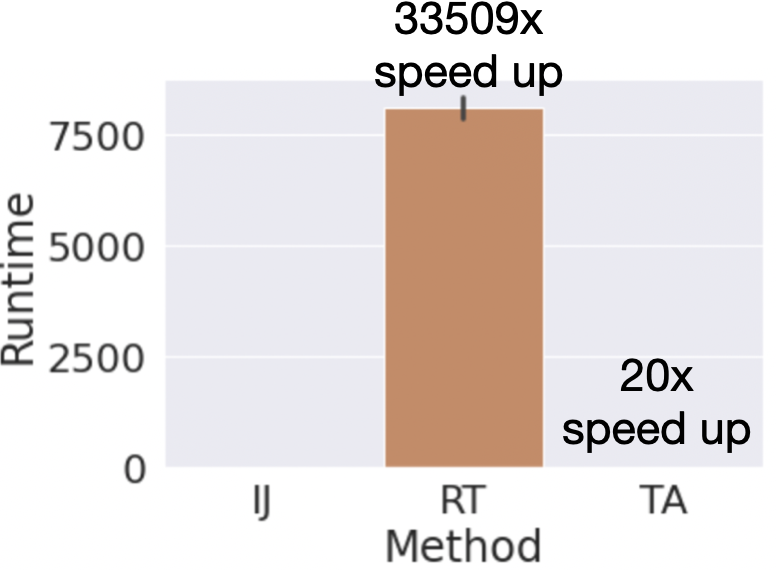}
  \label{fig:svhn_runtime_6}
\end{subfigure}%
\caption{\textbf{IJ} vs. \textbf{RT} vs. \textbf{TA} for non-convex settings on SVHN. Demonstrating runtime improvements across different hyperparameter settings of $10^{-4}, 10^{-5}, 10^{-6}$.}
\label{fig:svhn_runtimes}
\end{figure}

\begin{figure}[htb!]
\centering
\begin{subfigure}{.3\textwidth}
  \centering
  \includegraphics[width=\linewidth]{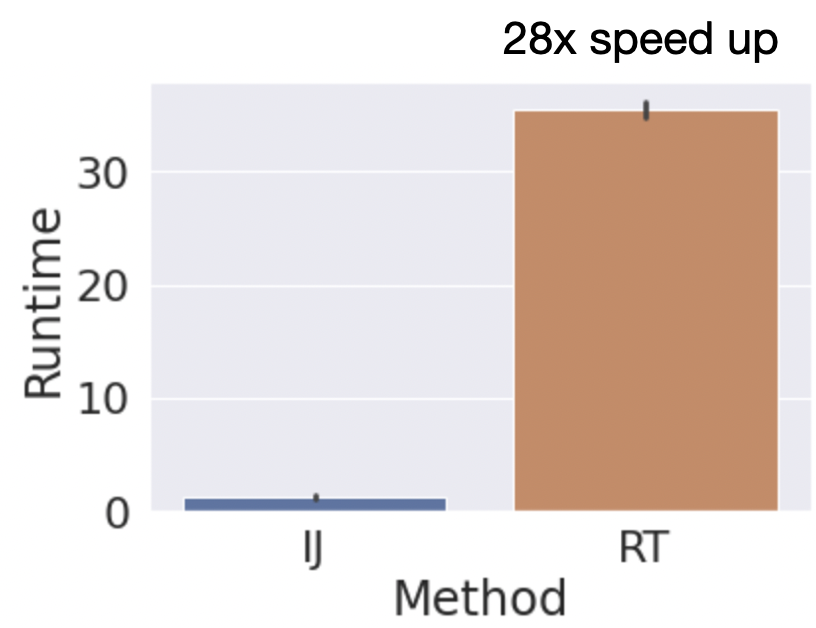}
  \label{fig:warfarin_runtime_4}
\end{subfigure}%
\begin{subfigure}{.3\textwidth}
  \centering
  \includegraphics[width=\linewidth]{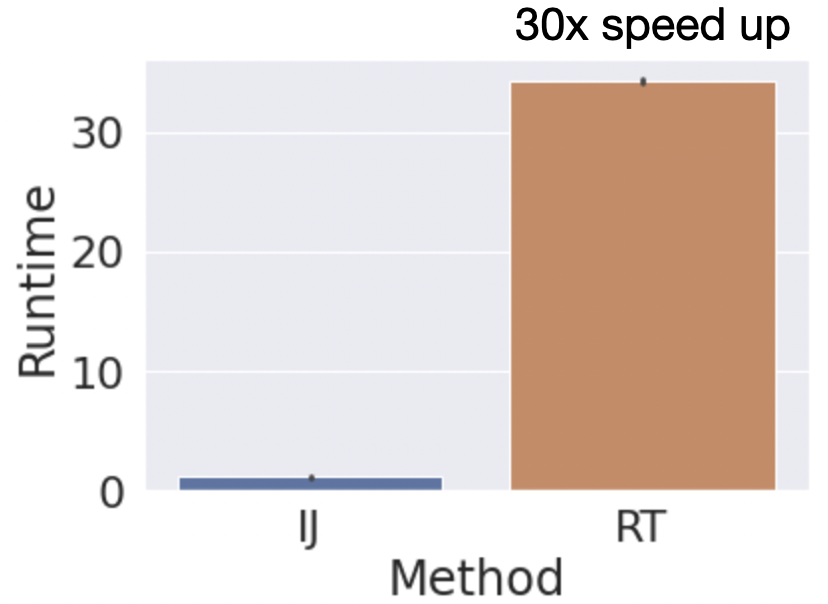}
  \label{fig:warfarin_runtime_5}
\end{subfigure}%
\begin{subfigure}{.3\textwidth}
  \centering
  \includegraphics[width=\linewidth]{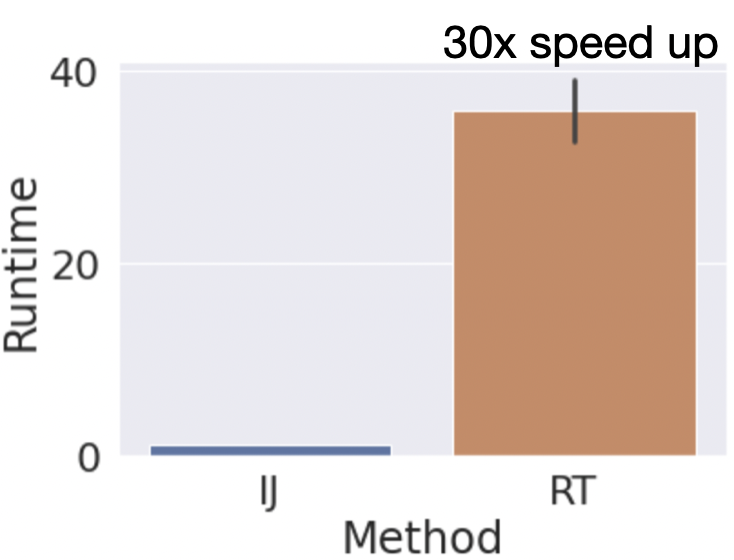}
  \label{fig:warfarin_runtime_6}
\end{subfigure}%
\caption{\textbf{IJ} vs. \textbf{RT} for non-smooth settings on Warfarin. Demonstrating runtime improvements across different hyperparameter settings of $10^{-4}, 10^{-5}, 10^{-6}$.}
\label{fig:warfarin_runtimes}
\end{figure}

% \end{proof}
% \end{lemma}

\end{document}